\documentclass[10pt,journal,compsoc]{IEEEtran}

\ifCLASSOPTIONcompsoc
  \usepackage[nocompress]{cite}
\else
  \usepackage{cite}
\fi

\usepackage{booktabs} % For formal tables
\usepackage{epsfig}
\usepackage{epstopdf}
\usepackage{multirow}
\usepackage{algorithm}
\usepackage{algorithmic}
\usepackage{amssymb}
\usepackage{amsfonts}
\usepackage{amsmath}
\usepackage{graphicx}
\usepackage{cancel}
\usepackage{diagbox}
\usepackage{color}
\usepackage{bm}
\usepackage{pifont}
\usepackage{tikz}
\usepackage{makecell}
\usepackage{multirow}
\usepackage{subfigure}
\usepackage{rotating}
\usepackage{hyperref}
\newcommand*{\circled}[1]{\lower.7ex\hbox{\tikz\draw (0pt, 0pt)%
    circle (.5em) node {\makebox[1em][c]{\small #1}};}}

\ifCLASSINFOpdf

\else

\fi

\begin{document}

%\title{Supervision Adaptation: Balancing In-Distribution Classification Generalization and Out-of-Distribution Detection}
%\title{Balancing In-Distribution Generalization and Out-of-Distribution Detection}
\title{Supervision Adaptation Balancing In-distribution Generalization and Out-of-distribution Detection}
%\title{Supervision Adaptation is All Out-of-Distribution Detection Needs.}

%~\IEEEmembership{Member,~IEEE,}
\author{Zhilin Zhao,
        Longbing Cao,
        and~Kun-Yu Lin% <-this % stops a space
\IEEEcompsocitemizethanks{\IEEEcompsocthanksitem Zhilin Zhao and Longbing Cao are with the Data Science Lab, School of Computing and DataX Research Centre, Macquarie University, Sydney, NSW 2109, Australia.\protect\\
E-mail: zhaozhl7@hotmail.com, longbing.cao@mq.edu.au
\IEEEcompsocthanksitem Kun-Yu Lin is with the School of Computer Science and Engineering, Sun Yat-sen University, Guangzhou, 515000, China.\protect\\
E-mail: kunyulin14@outlook.com
}% <-this % stops an unwanted space
}

\markboth{}%
{Shell \MakeLowercase{\textit{et al.}}: Bare Demo of IEEEtran.cls for Computer Society Journals}

\IEEEtitleabstractindextext{%
\begin{abstract}
The discrepancy between in-distribution (ID) and out-of-distribution (OOD) samples can lead to \textit{distributional vulnerability} in deep neural networks, which can subsequently lead to high-confidence predictions for OOD samples. This is mainly due to the absence of OOD samples during training, which fails to constrain the network properly. To tackle this issue, several state-of-the-art methods include adding extra OOD samples to training and assign them with manually-defined labels. However, this practice can introduce unreliable labeling, negatively affecting ID classification. The distributional vulnerability presents a critical challenge for non-IID deep learning, which aims for OOD-tolerant ID classification by balancing ID generalization and OOD detection. In this paper, we introduce a novel \textit{supervision adaptation} approach to generate adaptive supervision information for OOD samples, making them more compatible with ID samples. Firstly, we measure the dependency between ID samples and their labels using mutual information, revealing that the supervision information can be represented in terms of negative probabilities across all classes. Secondly, we investigate data correlations between ID and OOD samples by solving a series of binary regression problems, with the goal of refining the supervision information for more distinctly separable ID classes. Our extensive experiments on four advanced network architectures, two ID datasets, and eleven diversified OOD datasets demonstrate the efficacy of our supervision adaptation approach in improving both ID classification and OOD detection capabilities.
\end{abstract}

}

\maketitle

\IEEEdisplaynontitleabstractindextext

% \fi
\IEEEpeerreviewmaketitle

\IEEEraisesectionheading{\section{Introduction}\label{sec:introduction}}
Deep neural networks (DNNs) trained on independent and identically distributed (IID) samples from an unknown in-distribution (ID) demonstrate strong generalization when tested on samples from the same distribution~\cite{GE:17}. However, in real-world applications, test samples may incorporate changes from ID samples or originate from different, out-of-distribution (OOD) sources. A known issue with ID-trained networks is their tendency to make high-confidence, yet incorrect, predictions on these OOD samples~\cite{CAL:17}. This scenario partly falls under the general challenges of \textit{non-IID learning} \cite{C22beyiid,CaoY022} for shallow and deep models and more specifically causes the \textit{distributional vulnerability} \cite{Z22ood} of DNNs and other serious issues~\cite{UN:17}. For example, a security authentication system might be easily compromised by artificially generated samples~\cite{AD:15}, self-driving cars could make faulty decisions in new, uncertain, or complex situations which typically require expert judgment. The root cause of this overconfidence is the absence of OOD samples during the training phase, which fails to constrain the performance of DNNs on such samples. This leads to uncertain predictions for unknown OOD samples~\cite{DP:16} and in some cases, dangerous high-confidence predictions. Therefore, enhancing both ID generalization and OOD sensitivity during the training of DNNs is crucial for balanced and more reliable decision-making, mitigating the risk of disastrous outcomes.

A straightforward approach to mitigate the overconfidence issue is to introduce extra, label-free samples from different datasets as OOD samples to regularize the learning behavior of networks. These OOD samples usually receive manually-defined supervision information based on prior knowledge for training. However, this method compromises classification accuracy because the user-defined supervision for OOD samples is not coordinated with ID samples. Specifically, OOD-specific supervision information could disrupt the ID classification process. For instance, one common strategy for generating artificial supervision labels treats all OOD samples, regardless of their originating distribution, as members of a singular, extra class~\cite{OS:16}. This approach has its drawbacks. It creates an additional hyperplane to misclassify ID samples situated between different OOD samples in the data space as OOD, thereby reducing classification accuracy. Another technique generates a class probability vector from a flat distribution~\cite{PN:18}. This vector is linked to an OOD sample through an additional regularizer without altering the number of classes. This method could inadvertently skew the class distribution due to the introduction of these artificial class probability vectors. Moreover, because this regularizer is independent of ID samples, it may dilute the focus of networks on learning from the ID samples, further impacting classification capabilities.

The preceding discussion highlights that manually-assigned labels for additional OOD samples may not accurately capture their relationship with ID samples and can disrupt the ID classification learning process. A primary challenge is to provide adaptive supervision information for effectively integrating OOD samples with ID samples. This aims to achieve robust classification while also ensuring effective OOD detection. Such adaptive supervision information should fulfill two core objectives: distinguishing OOD samples from ID samples, and enhancing the balance between ID generalization and OOD detection capabilities. To realize these goals, adaptive supervision information should align OOD samples with specific regions in the data space that are distinct from areas occupied by ID samples. This improves the separability of ID samples that have different labels~\cite{GE:03}. Since the adaptive supervision information is to make OOD samples compatible with ID samples, it is crucial to examine the relationships both within the set of ID samples and between the sets of ID and OOD samples. These dual purposes allow adaptive supervision information to address two intertwined challenges: (1) integrating OOD samples with ID samples, and (2) enhancing the distinguishability among ID samples with different labels.

Accordingly, this paper addresses the aforementioned challenges by introducing a \textit{supervision adaptation} (SA) method, which is designed to harmonize the network generalization on ID samples with the detection capacity for OOD samples. It is worth noting that the conventional cross-entropy loss for ID samples originated from maximizing mutual information~\cite{VB:19} assumes that the data space contains only ID samples. To delve into the areas uncovered by ID samples, this study extends mutual information maximization to a mixed data space that includes both ID and OOD samples. A lower bound on the mutual information is derived to gauge the dependency between ID samples and their labels in this mixed space. Moreover, the study replaces an intractable conditional distribution within this bound with a solvable optimization problem, involving a parameterized discriminator. This lower bound indicates that supervision information for OOD samples could take the form of negative probabilities for all classes. To estimate this supervision information, the study examines data relationships between ID and OOD samples, aiming to further boost generalization capabilities. Guided by the core idea~\cite{GE:03} of making different ID classes more distinguishable, an OOD dataset is employed as a reference point. Specifically, a binary regression problem is solved to separate OOD samples from a given class of ID samples. Finally, a compact objective function for the SA method is derived by simplifying the combined results of both the lower bound on mutual information and the estimated supervision information for OOD samples.

In summary, SA tackles the aforementioned challenges through several innovative and crucial design elements:
\begin{itemize}
  \item To enhance the network sensitivity to OOD samples while minimizing their disruptive impact on ID classification, we unveil the form of adaptive supervision information suitable for OOD samples. This is accomplished by assessing the relationship between ID samples and their corresponding labels in a data space mixing both ID and OOD samples. The relationship is quantified using the lower bound of mutual information.
  \item To further boost classification accuracy, the SA method estimates adaptive supervision information by examining the relationships between OOD and ID samples that belong to the same class. This goal is realized by solving multiple binary regression problems.
  \item To achieve a balanced network performance in terms of generalization on ID samples and detection capability on OOD samples, SA combines the lower bound on mutual information in the mixed data space with the estimated supervision information for OOD samples. The combined results are then simplified to form a compact objective function.
\end{itemize}

In the following sections, the paper is organized as follows: Section~\ref{sec:relatedwork} provides a review of relevant literature. The SA method is elaborated in Section~\ref{sec:algorithm}. Section~\ref{sec:setup} outlines the experimental setups, while Section~\ref{sec:comRes} presents comparative results. Quantitative and qualitative analyses are covered in Section~\ref{sec:QA}. Finally, Section~\ref{sec:conclusion} summarizes our contributions and suggests directions for future research.

\section{Related Work}\label{sec:relatedwork}
We briefly review the methods for OOD detection~\cite{GOOD:21, OOD:21}, OOD generalization and domain adaptation~\cite{GUD:21,DG:23}, and generalization improvement~\cite{ML:14}.

\subsection{Out-of-distribution Detection}
A baseline method for detecting OOD samples employs a pretrained model that is optimized using cross-entropy loss on an ID training set~\cite{BL:17}. This method differentiates ID from OOD samples based on the maximum probabilities calculated from the softmax outputs. If the maximum probability of a sample surpasses a pre-established threshold, it is classified as ID; otherwise, it is deemed as OOD. Deep residual flow (DRF) \cite{DRF:20} uses a residual flow, a principled extension of Gaussian distribution modeling through non-linear normalizing flows, to identify OOD samples. Energy-based detector (EBD) \cite{EB:20} computes an OOD score for a test sample using an energy function, which is derived from the softmax activation of predicted label probabilities. Despite these innovations, the aforementioned methods primarily focus on developing OOD detectors to calculate OOD scores for existing pretrained networks without modifying them. This implies that the original sensitivity of the pretrained networks to OOD samples remains unaltered~\cite{ODIN:18}, as these networks are not fine-tuned to limit their behavior when encountering OOD samples.

An advanced method seeks to enhance the network sensitivity to OOD samples by incorporating extra samples that are markedly different from the ID training samples during the retraining process. For instance, a simple approach assigns an extra class (EC) to all the newly introduced OOD samples~\cite{OS:16}. Although this strategy can enhance OOD sensitivity, it has the drawback of making different OOD samples indistinguishable from ID samples, leading the network to misclassify some ID samples as OOD. Another technique involves pairing each OOD sample with an artificial probability vector through an additional regularization term, while keeping the number of classes unchanged. One variant employs cross-entropy loss for ID samples alongside a Kullback-Leibler (KL) divergence term, forcing the predicted distribution of OOD samples to approximate a uniform distribution (UD) \cite{GO:18}. Prior network (PN) \cite{PN:18} takes a similar approach but sets a dense Dirichlet distribution~\cite{LDA:03} as the target in the KL divergence. Rather than explicitly specifying a target for OOD samples, the outlier exposure (OE) method~\cite{OE:19} uses a margin ranking loss to increase the gap between the log probabilities of OOD and randomly-selected ID samples. Adversarial confidence enhancing training (ACET) \cite{ADB:19} explores the neighborhood of OOD samples and imposes low confidence on these areas. The ensemble of self-supervised leave-out classifiers (ESLC) \cite{ESS:18} uses a random subset of training samples as OOD while treating the rest as ID, employing real-world OOD samples only during the validation process to select and integrate multiple OOD-sensitive networks. Nonetheless, these methods, which rely on artificially-labeled OOD samples, may misguide networks due to the alteration of label distributions. The regularization techniques employed do not contribute to the classification of ID samples, but divert the focus of networks away from this essential task.

Other innovative approaches focus on improving OOD sensitivity by altering loss functions and training protocols, especially when training OOD samples are not available. One such method is the generalized out-of-distribution image detection in neural networks (GODIN)~\cite{DCC:20}. It adaptively learns temperature scaling and perturbation magnitude, constructing an objective function by decomposing the posterior distribution. Another approach, known as minimum other score (MOS)~\cite{MOS:21}, is based on the observation that increasing the number of ID classes tends to reduce the performance of OOD detection. In response, MOS groups classes with similar concepts to effectively lower the number of ID classes, aiming to improve OOD sensitivity. Self-supervised outlier detection (SSD) \cite{SSD:21} employs contrastive learning with image transformation, considering the Mahalanobis distance to the closest class centroid as the OOD score. Fine-tuning explicit discriminators by implicit generators (FIG) \cite{Z22ood} introduces an implicit generator for a pretrained network. This implicit generator creates specific OOD samples designed to yield low-confidence predictions, thereby increasing the network sensitivity to OOD samples. However, these methods come at a cost: they often sacrifice the generalization ability of the network on ID samples, which may make them less applicable in real-world settings where both high ID generalization and OOD sensitivity are crucial.

\subsection{Out-of-distribution Generalization and Domain Adaptation}
OOD generalization aims to achieve robust performance on OOD samples that exhibit covariate shifts, relying solely on training data from the source domain. Notably, this differs from OOD detection, where the focus is on identifying samples that diverge semantically from the ID data. Representation self-challenging (RSC)~\cite{RSC:20} nudges neural networks to discover less prominent, yet meaningful features by systematically sidelining dominant features present in the training data. Invariant risk minimization (IRM)~\cite{IRM:19} harnesses nonlinear, invariant, and causal predictors from multiple training conditions to bolster OOD generalization capabilities. ANDMask~\cite{AND:21} mitigates the risk of model memorization by nullifying gradient components tied to inconsistently weighted features across different environments. Group distributionally robust optimization (GroupDRO)~\cite{DRO:20} trains the model to excel under worst-case distribution scenarios, guided by explicit group annotations in the dataset. Kernelized heterogeneous risk minimization (KerHRM)~\cite{KHRM:21} engages in both latent heterogeneity exploration and invariant prediction, utilizing an orthogonal, heterogeneity-aware kernel and invariant gradient descent. Stable adversarial learning (SAL)~\cite{SAL:21} focuses on optimizing against worst-case costs within a realistically-defined set of uncertain covariates. Distributionally invariant learning (DIL)~\cite{DIL:22} loosens the stringent requirements for invariance across multiple environments by iteratively identifying and addressing the worst-performing sub-populations. Gradient norm aware minimization (GAM)~\cite{GNA:23} refines the model generalization by pursuing minima that exhibits uniformly low curvature in all directions.

When unlabeled samples from the target domain become accessible, the challenge of OOD generalization transits into the realm of domain adaptation~\cite{CI:22}. In this context, the objective is to leverage insights from a source domain to attain robust generalization capabilities for a closely related target domain. The Kullback-Leibler importance estimation procedure (KLIEP) \cite{KLIEP:07} aims to minimize the KL-divergence between the genuine target distribution and a modified, importance-weighted source distribution. Similarly, domain adaptation support vector machine (DASVM) \cite{DASVM:10} iteratively integrates labeled samples from the target domain into the training set, modulating the cost factors for errors in the target and source domains dynamically. The domain-adversarial neural network (DANN) \cite{DANN:16} seeks to identify a feature space where the source and target domains become indistinguishable, while maintaining accurate classification on the source domain. Lastly, Deep correlation alignment (DCORAL) \cite{CORAL:16} adjusts the correlation structures between layer activations in DNNs to mitigate the effects of domain shift.

\subsection{Generalization Improvement}
Data augmentation methods can enhance the ability of a network to generalize when classifying ID samples. They often produce variants that closely resemble OOD samples~\cite{DA:12}. These techniques are particularly instrumental in mitigating overfitting by training on perturbed yet relevant instances. Widely utilized data augmentation strategies in computer vision encompass rotation, translation, cropping, resizing, flipping~\cite{FL:15}, and random erasing~\cite{RE:20}. A straightforward approach involves the addition of Gaussian noise (GN) \cite{GN:17} to samples, treating these noise-injected variants as OOD instances. However, such data-independent perturbations may lack informative cues for effective OOD detection. In a more nuanced strategy, MIXUP \cite{MIXUP:18} generates virtual training samples through convex combinations of random pairings of samples and their one-hot label vectors. When the combined samples derive from different classes with nearly equal weights, these generated instances can be considered OOD, owing to their minimal target class probabilities. Separately, certain approaches focus solely on bolstering generalization by manipulating the label vectors of ID samples. For example, label smoothing (LS) \cite{CP:17} dilutes a one-hot label vector by blending it with a uniformly-distributed random vector, while teacher-free knowledge distillation (Tf-KD) \cite{KD:20} combines this one-hot vector with the output from a pretrained teacher network. Although these techniques may improve ID classification, they often hinder the ability of a network to effectively distinguish OOD samples. The core issue lies in the narrowing of the confidence gap between ID and OOD samples, as reducing confidence on ID instances without simultaneously constraining predictions for OOD samples can inadvertently normalize the latter, making them harder to detect as outliers.

\begin{figure}
  \centering
  \includegraphics[width=0.45\textwidth]{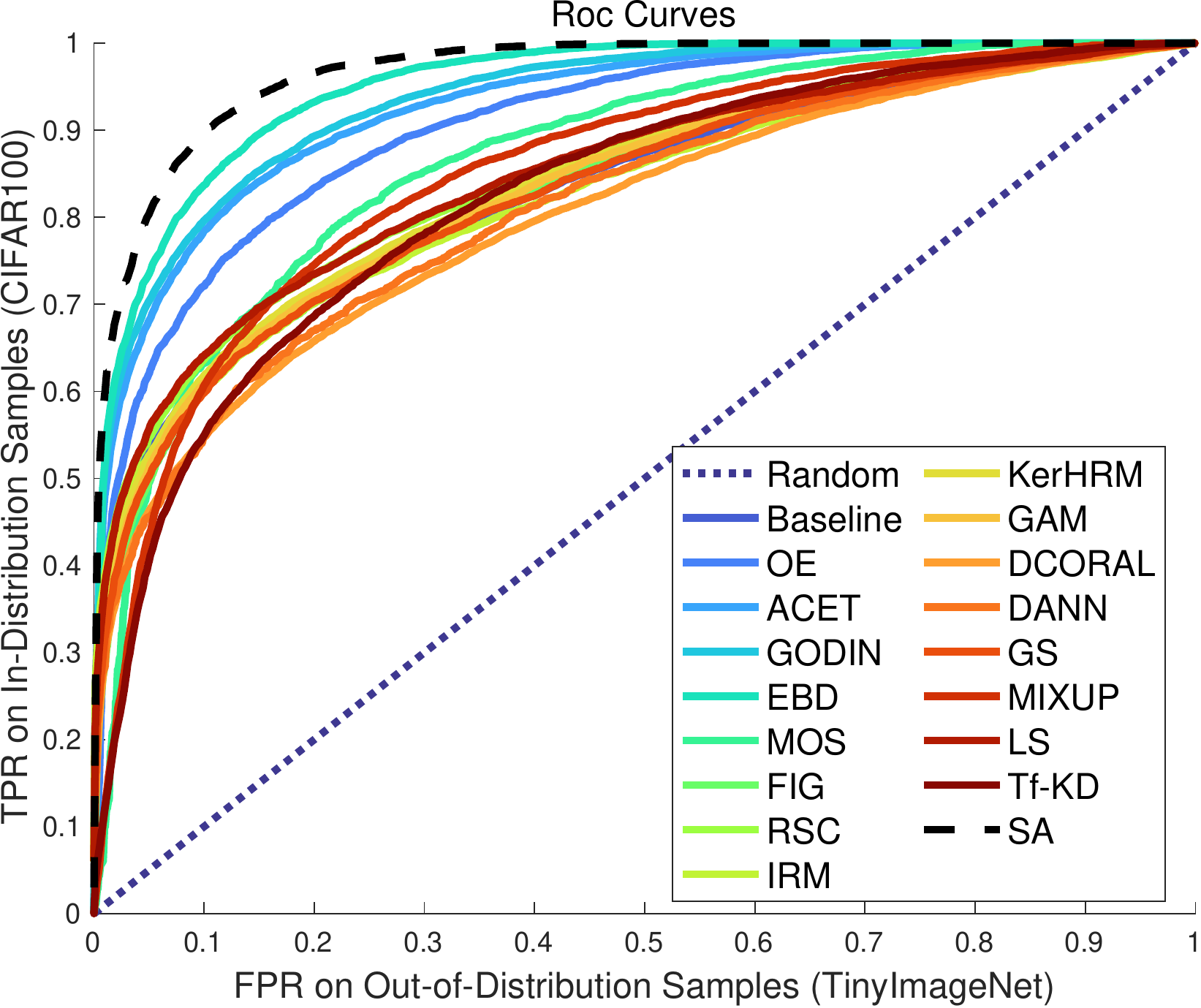}
  \caption{The ROC curve on ResNet18 and CIFAR100. The dashed, dotted, and solid curves correspond to the ROC curves of the proposed SA method, the random method, and other compared methods, respectively. The random method assigns the confidence scores drawn from a uniform distribution to any samples. The corresponding test set of CIFAR100 is used to calculate the TPR values, and the rest OOD samples from TinyImageNet after training SA are used to calculate the FPR values. The area under the ROC curve represents the AUROC score, where a larger score indicates the better OOD detection performance.}
  \label{fig:curve}
\end{figure}

In comparison, \figurename~\ref{fig:curve} illustrates the performance of OOD detection by the SA method and state-of-the-art alternatives, quantified by the area under the precision-recall curve (AUROC). The detailed discussion regarding the experimental setup and comparative results can be found in Sections~\ref{sec:setup} and~\ref{sec:comRes}, respectively.

\section{Supervision Adaptation}\label{sec:algorithm}
We present our \textit{supervision adaptation} (SA) method, a novel approach designed to achieve a balance in performance between ID generalization and OOD detection. SA constructs adaptive supervision information for OOD samples to improve OOD detection without compromising ID classification performance. The objective function of SA incorporates two key components: Mixed Space Mutual Information (MSMI) and Multiple Binary Cross-Entropy (MBCE). MSMI defines the type of supervision information for an OOD sample by utilizing negative probabilities from all classes. This allows the model to gain a nuanced understanding of how the OOD sample deviates from known distributions. On the other hand, MBCE plays a crucial role in estimating this supervision information. It also dictates the formulation of the parameterized discriminator within the MSMI framework. The final objective function for SA is derived by calculating the lower bound of the aggregated outcomes from both MSMI and MBCE. This multi-faceted objective function enables SA to provide a balanced and effective solution for both ID generalization and OOD detection.

\subsection{Problem Statement}
Let us consider a feature vector $\mathbf{x}$ and a corresponding label $y$. We denote the distributions for ID and OOD samples as $P_I(\mathbf{x})$ and $P_O(\mathbf{x})$, respectively. We can describe a mixed distribution of these samples as follows:
\begin{equation}
P(\mathbf{x}) = (1 - \epsilon) P_I(\mathbf{x}) + \epsilon P_O(\mathbf{x})
\end{equation}
where $\epsilon$ is a component parameter that controls the proportions of ID and OOD samples in the mixed data space. Note that only ID samples possess ground-truth labels, and introducing label-free OOD samples should not alter the existing label distribution. We also make two critical assumptions: (1) the label distributions for ID samples and the mixed data set remain identical, i.e., $P_I(y) = P(y)$, and (2) the joint distributions of the feature-label pair $(\mathbf{x}, y)$ for ID and mixed data are congruent, i.e., $P_I(\mathbf{x},y) = P(\mathbf{x},y)$. Subsequently, the marginal distribution $P_I(\mathbf{x})$ can be expressed as:
\begin{equation}
\begin{aligned}
P_I(\mathbf{x}) & = \sum_{y \in [K]} P_I(\mathbf{x},y)  = \sum_{y \in [K]} P_I(y|\mathbf{x}) P(y) \\
& = \sum_{y \in [K]} P_I(y|\mathbf{x}) P_I(y),
\end{aligned}
\end{equation}
where $K$ signifies the total number of classes.

In the context of SA, we aim to train a parameterized discriminator $Q_{\theta}(y|\mathbf{x})$ utilizing an ID dataset $\mathcal{D}^I = \{\mathbf{x}^I_i, y^I_i\}_{i = 1}^M$ and an OOD dataset $\mathcal{D}^O = \{\mathbf{x}^O_i\}_{i = 1}^N$. Here, $\theta$ denotes the model parameter, and the ratio $M / N = (1 - \epsilon) / \epsilon$. The central research question we aim to address is: The central research question we aim to address is twofold. First, given a sample originating from the mixed distribution $P(\mathbf{x})$, can the discriminator $Q_{\theta}(y|\mathbf{x})$ effectively determine whether the sample is drawn from $P_O(\mathbf{x})$? Second, if the sample stems from $P_I(\mathbf{x})$, can the discriminator accurately assign an appropriate label to it?

\subsection{MSMI: Mixed Space Mutual Information}
We characterize the supervision information by examining the relationship between samples and labels within the mixed data space. Our core assumption is that \textit{an ID sample has a strong association with its specific label, while an OOD sample has a weaker connection to any class}. Due to the weak association between OOD samples and classes, we opt out to explicitly define the supervision information for OOD samples, as is commonly done in other works. Instead, we implicitly deduce this information by amplifying the relationship between ID samples and their corresponding labels within the mixed data space. This approach not only makes the network more sensitive to OOD samples but also minimizes their interference when classifying ID samples.

Mutual information (MI) quantifies the interdependence between random variables. Traditional approaches utilizing MI~\cite{MINE:18} often assume that the data space solely consists of ID samples, an assumption that does not account for the presence of OOD samples. To address this limitation, we extend MI into a mixed data space and capture the relationship between ID samples and their corresponding labels as follows:
\begin{equation}
\begin{aligned}
\mathcal{I}(X;Y) & = \mathbb{E}_{P(\mathbf{x},y)} \left[\log \frac{P(\mathbf{x},y)}{P(\mathbf{x})P(y)}\right] \\
& = \mathbb{E}_{P_I(\mathbf{x},y)} \left[\log \frac{P_I(\mathbf{x},y)}{P(\mathbf{x})P_I(y)}\right].
\label{eqn:MI}
\end{aligned}
\end{equation}
Unlike traditional MI methods that presume OOD samples to follow the distribution of ID samples, the distribution $P(\mathbf{x},y)$ in our mixed data space incorporates both ID and OOD samples. Since ground-truth labels are only available for ID samples, Eq.~(\ref{eqn:MI}) explicitly measures the relationship between ID samples and their labels. Nevertheless, by optimizing Eq.~(\ref{eqn:MI}) over the mixed distribution $P(\mathbf{x})$, we can indirectly infer the supervision information for OOD samples, thereby facilitating their adaptation to ID samples.

In the training process, observed samples are drawn from the mixture distribution $P(\mathbf{x})$. These samples are associated with a label according to the conditional distribution $P(y|\mathbf{x})$ if they are ID samples, or remain label-free if they are OOD samples. Guided by the mixed space MI framework, our goal is to estimate $P(y|\mathbf{x})$ by training a parameterized discriminator $Q_{\theta}(y|\mathbf{x})$ to maximize Eq.~(\ref{eqn:MI}). Though computing MI directly is challenging due to our lack of access to the true distributions, we can still compute gradients for a lower bound~\cite{LB:03} on MI with respect to the parameter $\theta$. To derive this lower bound, we decompose MI into tractable terms and employ $Q_{\theta}(y|\mathbf{x})$ to approximate the unknown conditional distribution $P(y|\mathbf{x})$, i.e.,
\begin{equation}
\begin{aligned}
\label{eq:info}
\mathcal{I}(X;Y) & = \mathbb{E}_{P_I(\mathbf{x},y)} \left[\log \frac{P_I(\mathbf{x},y)}{P(\mathbf{x})}\right] - \mathbb{E}_{P_I(\mathbf{x})} \mathbb{E}_{P_I(y)} \left[\log P_I(y) \right]\\
& = \mathbb{E}_{P_I(\mathbf{x},y)} \left[\log \frac{P_I(\mathbf{x},y)}{P(\mathbf{x})} - \log P_I(y)\right] \\
& = \mathbb{E}_{P_I(\mathbf{x},y)} \left[\log \frac{P_I(\mathbf{x},y)}{P(\mathbf{x})}\right] + H(y)\\
& = \mathbb{E}_{P_I(\mathbf{x},y)} \left[\log \frac{P_I(\mathbf{x},y) Q_{\theta}(y|\mathbf{x}) }{P(\mathbf{x}) Q_{\theta}(y|\mathbf{x})}\right] + H(y),\\
\end{aligned}
\end{equation}
where $H(y) = - \mathbb{E}_{P_I(y)} \left[\log P_I(y) \right] \in [0, \log K]$, referring to the entropy of variable $y$. Using Bayes's theorem and the assumption $P(\mathbf{x},y) = P_I(\mathbf{x},y)$, we have
\begin{equation}
\begin{aligned}
\label{eq:baye}
P(y | \mathbf{x}) = \frac{P(\mathbf{x},y)}{P(\mathbf{x})} = \frac{P_I(\mathbf{x},y)}{P(\mathbf{x})}.
\end{aligned}
\end{equation}
Accordingly, we obtain
\begin{equation}
\begin{aligned}
\label{eq:t2}
& \mathbb{E}_{P_I(\mathbf{x},y)} \left[\log \frac{P_I(\mathbf{x},y) }{P(\mathbf{x}) Q_{\theta}(y|\mathbf{x})}\right]\\
= & \mathbb{E}_{P(y | \mathbf{x}) P(\mathbf{x})} \left[\log \frac{P(y|\mathbf{x}) }{Q_{\theta}(y|\mathbf{x})}\right]\\
= & \mathbb{E}_{P(\mathbf{x})} \left[ D_{\text{KL}} \left( P(y|\mathbf{x}) \| Q_{\theta}(y|\mathbf{x})\right) \right].\\
\end{aligned}
\end{equation}
Due to the non-negativity of entropy $H(y)$ and Eq.~(\ref{eq:t2}), we can obtain the lower bound of Eq.~(\ref{eq:info}),
\begin{equation}
\begin{aligned}
\mathcal{I}(X;Y) \geq & \mathbb{E}_{P_I(\mathbf{x},y)}\left[ \log Q_{\theta}(y|\mathbf{x}) \right]  \\
& + \mathbb{E}_{P(\mathbf{x})} \left[ D_{\text{KL}} \left( P(y|\mathbf{x}) \| Q_{\theta}(y|\mathbf{x})\right) \right].\\
\end{aligned}
\end{equation}
Here, the last equality is attributed to the KL-divergence. By retaining the terms relating to $Q_{\theta}(y|\mathbf{x})$, we can obtain the objective function of MSMI,
\begin{equation}
\begin{aligned}
\label{eq:msmi}
\mathcal{L}_{\text{MSMI}}(\theta) = & \mathbb{E}_{P_I(\mathbf{x},y)}\left[ \log Q_{\theta}(y|\mathbf{x}) \right]  \\
& + \beta \mathbb{E}_{P(\mathbf{x})} \left[ \sum_{y \in [K]} -P(y|\mathbf{x}) \log Q_{\theta}(y|\mathbf{x}) \right],\\
\end{aligned}
\end{equation}
where $\beta$ serves as a hyper-parameter controlling the degree of regularization on the outputs for ID and OOD samples. The first term in the objective function relies on the ground-truth labels of ID samples. The second term, however, encapsulates supervision information for both ID and OOD samples in the form of negative log probabilities across all $K$ classes:
\begin{equation}
[-P(1|\mathbf{x}),\ldots,-P(y|\mathbf{x}),\ldots,-P(K|\mathbf{x})].
\end{equation}
For ID samples, this approach utilizes two types of supervision information: the ground-truth labels and the negative probabilities of all classes. For OOD samples, the negative probabilities of all classes act as adaptive supervision information. Importantly, this probabilistic supervision also constrains the behavior of $Q_{\theta}(y|\mathbf{x})$ for ID samples without negatively affecting their classification. This is evident when the data space contains only ID samples, reducing Eq.~(\ref{eq:msmi}) to the confidence penalty \cite{CP:17}, an approach empirically shown to enhance generalization.

\subsection{MBCE: Multiple Binary Cross Entropy}
As per the design outlined in Eq.~(\ref{eq:msmi}), the supervision information $[-P(1|\mathbf{x}),\ldots,-P(y|\mathbf{x}),\ldots,-P(K|\mathbf{x})]$ serves to adapt OOD samples to ID samples. This is accomplished by reinforcing the association between ID samples and their corresponding labels, thereby enhancing the awareness of networks of OOD samples and reducing their interference during the classification of ID samples. This design choice stems from maximizing the mutual information between ID samples and their labels in a mixed data space, as described by Eq.~(\ref{eq:info}). However, it is worth noting that directly providing such supervision information for OOD samples during training is impractical, largely because the conditional distribution $P(y|\mathbf{x})$ for OOD samples is typically unknown. To circumvent this challenge, we estimate this supervision information by investigating the data relationships between ID and OOD samples. Specifically, we make the reasonable assumption that OOD samples are distinct from ID samples in their affiliation with specific classes. This distinction serves to further separate ID samples associated with different labels, thereby making the classification task more robust against OOD interference.

It is important to remember that the parametric discriminator $Q_{\theta}(y|\mathbf{x})$ is used to approximate $P(y|\mathbf{x})$ for label prediction. Consequently, $Q_{\theta}(y|\mathbf{x})$ can be modified to encapsulate adaptive supervision information specifically for OOD samples. For instance, conventional approaches often employ the standard softmax function on $Q_{\theta}(y|\mathbf{x})$ to estimate $P_I(y|\mathbf{x})$. However, this method falls short in approximating $P(y|\mathbf{x})$ as it neglects the data relationships between ID and OOD samples. This omission fails to adapt OOD samples to ID ones, thereby potentially skewing the learning process for label prediction on ID samples.

To enhance ID classification accuracy by effectively separating different classes of ID samples~\cite{RC:02}, we tailor $Q_{\theta}(y|\mathbf{x})$ to both estimate the supervision information for OOD samples and approximate the conditional distribution in the MI framework within the mixed data space. Specifically, we construct $K$ binary regression problems, where $K$ represents the number of classes. Each binary regression is designed to differentiate between ID samples of a specific label and all OOD samples. Subsequently, we synthesize these $K$ individual problems to arrive at a unified expression for $Q_{\theta}(y|\mathbf{x})$.

To estimate the conditional distribution $P(y|\mathbf{x})$, we employ the parameterized discriminator $Q_{\theta}(y|\mathbf{x})$,
\begin{equation}
\begin{aligned}
\label{eq:err}
Q_{\theta}(y | \mathbf{x}) = \frac{P_I(\mathbf{x},y)}{P(\mathbf{x})} =  \frac{P_I(\mathbf{x},y)}{(1 - \epsilon) \sum_{y \in [K]} P_I(\mathbf{x},y) + \epsilon P_O(\mathbf{x})}. \\
\end{aligned}
\end{equation}
However, designing $Q_{\theta}(y | \mathbf{x})$ to approximate $P(y|\mathbf{x})$ according to Eq.~(\ref{eq:err}) is problematic, as it necessitates the estimation of $P(\mathbf{x})$. This in turn requires $P_I(\mathbf{x},y)$, which is unknown for OOD samples. To circumvent this issue, we initially reformulate Eq.~(\ref{eq:err}),
\begin{equation}
\begin{aligned}
\label{eq:Q1}
Q_{\theta}(y | \mathbf{x}) = \frac{P_I(\mathbf{x},y)}{P_I(\mathbf{x},y) + P(\mathbf{x})} \times \frac{P_I(\mathbf{x},y) + P(\mathbf{x})}{P(\mathbf{x})},\\
\end{aligned}
\end{equation}
and estimate the following density ratio~\cite{DE:12} by an auxiliary function $\mathcal{D}(\mathbf{x},y)$,
\begin{equation}
\begin{aligned}
\label{eq:Q2}
\frac{P_I(\mathbf{x},y)}{P(\mathbf{x}) + P_I(\mathbf{x},y)} = \mathcal{D}(\mathbf{x},y) =  \sigma (\log \varphi(\mathbf{x},y) ), \\
\end{aligned}
\end{equation}
where
\begin{equation}
\sigma(s) = 1 / (1 + \exp(-s)),
\end{equation}
is the sigma function and $\varphi(\mathbf{x},y)$ is the normalized output by the softmax function,
\begin{equation}
\begin{aligned}
\label{eq:sm}
\varphi(\mathbf{x},y) = \frac{ \exp\left(f_{\theta}(\mathbf{x},y)\right) }{ \sum_{y \in [K]}  \exp\left( f_{\theta}(\mathbf{x},y)\right) },\\
\end{aligned}
\end{equation}
where $f_{\theta}$ is a parametric neural network which maps each sample $\mathbf{x}$ to a \textit{K}-dimensional output vector $(f_{\theta}(\mathbf{x},1),\ldots,f_{\theta}(\mathbf{x},y),\ldots,f_{\theta}(\mathbf{x},K))$, and each $f_{\theta}(\mathbf{x},y)$ represents the classification score of the corresponding class.

Substituting Eq.~(\ref{eq:Q2}) into Eq.~(\ref{eq:Q1}), we have,
\begin{equation}
\begin{aligned}
\label{eq:Q3}
Q_{\theta}(y | \mathbf{x})
= & \frac{\mathcal{D}(\mathbf{x},y)}{1 - \mathcal{D}(\mathbf{x},y)} = \exp\left( \log \frac{\mathcal{D}(\mathbf{x},y)}{1 - \mathcal{D}(\mathbf{x},y)} \right)\\
= & \exp\left(\sigma^{-1}(\mathcal{D}(\mathbf{x},y)) \right) = \exp\left( \sigma^{-1}(\sigma(\log \varphi(\mathbf{x},y))) \right)\\
= & \exp\left( \log \varphi(\mathbf{x},y)\right) = \varphi(\mathbf{x},y).
\end{aligned}
\end{equation}
The third equality of Eq.~(\ref{eq:Q3}) corresponds to the property about the sigma function,
\begin{equation}
\sigma^{-1}(s) = \log (s / (1 - s)).
\end{equation}
Therefore, we arrive at a concise yet nuanced expression $Q_{\theta}(y | \mathbf{x}) = \varphi(\mathbf{x},y)$. Although this formula may resemble traditional methods on the surface, its underlying interpretation is fundamentally different. Here, it serves as an intermediate step for solving the auxiliary function $\mathcal{D}(\mathbf{x},y)$ in Eq.~(\ref{eq:Q2}), as opposed to being a target in traditional objective functions such as $\max \mathbb{E}_{P_I(\mathbf{x},y)}\left[ \log \varphi(\mathbf{x},y) \right]$. In other words, solving $\mathcal{D}(\mathbf{x},y)$ provides a particular form of $Q_{\theta}(y | \mathbf{x})$, and this form is subsequently used in optimizing the objective function in Eq.~(\ref{eq:msmi}). This process aims to train $Q_{\theta}(y | \mathbf{x})$ such that it approximates $P(y|\mathbf{x})$ as closely as possible, thus effectively capturing the mutual information within a mixed data space of both ID and OOD samples.

The task of deciding the form of $Q_{\theta}(y | \mathbf{x})$ essentially boils down to solving $\mathcal{D}(\mathbf{x},y)$. Observing that $\mathcal{D}(\mathbf{x},y)$ essentially captures the likelihood of a sample
$\mathbf{x}$ coming from the joint distribution $P_I(\mathbf{x},y)$ relative to the mixture distribution $P(\mathbf{x})$, this problem is analogous to a binary classification problem as indicated by previous work~\cite{EIM:20}. Here, the samples originated from $P_I(\mathbf{x},y)$ are labeled positive, while those from $P(\mathbf{x})$ are labeled negative. A significant challenge arises when considering the label distribution for these negative samples, as the labels from OOD samples in the mixed data space are not known a priori. To circumvent this issue, we make an assumption that the label distribution $P(y)$ of the mixture distribution $P(\mathbf{x})$ aligns with that of $P_I(y)$. This ensures that the influence of OOD samples does not distort the ID sample classification process by skewing the known label distribution $P_I(y)$.

Thus, drawing upon the framework of logistic regression~\cite{PRML:06}, we train the density ratio $\mathcal{D}(\mathbf{x},y)$ by maximizing an objective function of MBCE,
\begin{equation}
\begin{aligned}
\label{eq:mbce}
\mathcal{L}_{\text{MBCE}}(\theta) = & \mathbb{E}_{P_I(\mathbf{x},y)} \left[ \log \mathcal{D}(\mathbf{x},y) \right]\\
&- \mathbb{E}_{P_I(y)}\mathbb{E}_{P(\mathbf{x})} \left[ \log \left( 1 - \mathcal{D}(\mathbf{x},y) ) \right)\right].
\end{aligned}
\end{equation}
This function would aim to distinguish between samples from $P_I(\mathbf{x},y)$ and $P(\mathbf{x})$ as effectively as possible, and its optimized value should provide a robust estimate for $\mathcal{D}(\mathbf{x},y)$. This, in turn, informs the choice of $Q_{\theta}(y | \mathbf{x})$, leading to a more accurate approximation of $P(y|\mathbf{x})$.

In order to gain a deeper understanding of this objective function, we rewrite the first term according to the product rule of probability and have,
\begin{align}
\mathbb{E}_{P_I(y)} \left[ \mathbb{E}_{P_I(\mathbf{x}|y)} \left[ \log \mathcal{D}(\mathbf{x},y) \right] - \mathbb{E}_{P(\mathbf{x})} \left[ \log \left( 1 - \mathcal{D}(\mathbf{x},y) ) \right)\right] \right].
\end{align}
The objective function Eq.~\ref{eq:mbce} encompasses an ensemble of multiple binary regression problems, each applying a binary cross-entropy loss. The number of these losses equates to the total count of in-distribution (ID) labels. Each loss serves to differentiate ID samples from their respective conditional distribution $P_I(\mathbf{x}|y)$ as against those from the global mixture distribution $P(\mathbf{x})$. Significantly, the OOD samples, which are common across these binary regression problems, act as a unifying element. They effectively bridge the gap between ID samples of varying labels, making them more discernible from one another. This bridging mechanism enhances the ability of networks to classify ID samples with greater accuracy. It is worth noting that all binary regression problems share the same parameterized neural network $f_\theta$, allowing us to cohesively optimize the binary cross-entropy losses via the integrated loss Eq.~(\ref{eq:mbce}). Upon optimizing this objective function, we extract the estimated supervision information $[-P(1|\mathbf{x}),\ldots,-P(y|\mathbf{x}),\ldots,-P(K|\mathbf{x})]$ for OOD samples, along with the expression $Q_{\theta}(y | \mathbf{x})$.

However, it is essential to highlight that although $Q_{\theta}(y | \mathbf{x})$ can be derived from optimizing $\mathcal{D}(\mathbf{x},y)$ in Eq.~(\ref{eq:mbce}), the true learning objective should be to maximize mutual information between ID samples and their labels, as stipulated in Eq.~(\ref{eq:msmi}). This is because Eq.~(\ref{eq:mbce}) serves as a constraint that captures the intricate data relationships between ID and OOD samples, which in turn influences $Q_{\theta}(y | \mathbf{x})$ as defined in Eq.~(\ref{eq:msmi}).

\begin{algorithm*}[t]
    \caption{Supervision Adaptation (SA)}
    \label{alg:SA}
    \begin{algorithmic}[1]
        \STATE Sample $\mathcal{D}^I = \{\mathbf{x}^I_i, y^I_i\}_{i = 1}^M$ from $P_I(\mathbf{x},y)$
    \STATE Sample $\mathcal{D}^O = \{\mathbf{x}^O_i\}_{i = 1}^N$ from $P_O(\mathbf{x})$
    \REPEAT
    \STATE Estimate objective function:
        \begin{align*}
       \mathcal{\widetilde{L}}_{SA}(\theta) = \frac{1}{M}\sum_{(\mathbf{x},y) \in \mathcal{D}^I } \log \varphi(\mathbf{x}, y) +  \frac{\alpha}{M + N}\sum_{\mathbf{x} \in \mathcal{D}^I \cup \mathcal{D}^O} \left[ \sum_{y \in [K]} \left(  P_I(y) - \varphi(\mathbf{x},y) \right) \log \varphi(\mathbf{x},y) \right]
        \end{align*}
    \STATE Obtain gradients $\nabla_{\theta} \mathcal{\widetilde{L}}_{SA}(\theta)$ to update parameters $\theta$
    \UNTIL{convergence}
    \STATE {\bfseries Output:} discriminator $Q_{\theta}(y|\mathbf{x}) =  \varphi(\mathbf{x},y) $
\end{algorithmic}
\end{algorithm*}

\subsection{Objective Function}
The proposed SA method aims to pursue a balance between ID generalization and OOD detection. To achieve this goal, SA leverages two key objective functions: MSMI and MBCE. MSMI serves to bring the awareness of networks to OOD samples while minimizing their negative influence on ID classification. Specifically, MSMI helps elucidate the required form of supervision information for effectively handling OOD samples. On the other hand, MBCE provides the mechanism for estimating this supervision information and determines the expression for the parameterized discriminator within the MSMI framework, thereby improving generalization on ID samples. In summary, MSMI and MBCE are complementary components that together form a more robust approach. By integrating these two objective functions, we derive a comprehensive objective function tailored for the SA method, thereby achieving an optimized balance between ID generalization and OOD detection.

To incorporate the adaptive supervision information $[-P(1|\mathbf{x}),\ldots,-P(y|\mathbf{x}),\ldots,-P(K|\mathbf{x})]$ for OOD samples into the training of the discriminator $Q_{\theta}(y | \mathbf{x})$, and to further specify the design of $Q_{\theta}(y | \mathbf{x})$ by approximating this supervision information, we adopt a linear combination of MSMI and MBCE. This is because MBCE effectively serves as a constraint on the parameterized discriminator $Q_{\theta}(y | \mathbf{x})$ within the MSMI framework. Accordingly, we have:
\begin{equation}
\begin{aligned}
\label{eq:SA1}
(1 - \alpha) \mathcal{L}_{\text{MSMI}}(\theta) + \alpha \mathcal{L}_{\text{MBCE}}(\theta),
\end{aligned}
\end{equation}
where $\alpha \in [0,1]$ serves as a combination parameter to balance the impact of the two components. Combining Eqs. (\ref{eq:Q1}), (\ref{eq:Q3}) and (\ref{eq:SA1}), we have,
\begin{equation}
\begin{aligned}
\label{eq:SA2}
& \underbrace{(1 - \alpha) \mathbb{E}_{P_I(\mathbf{x},y)} \left[ \log \varphi(\mathbf{x},y) \right]}_{:= A} \\
& \underbrace{-(1 - \alpha) \beta \mathbb{E}_{P(\mathbf{x})} \left[ \sum_{y} \varphi(\mathbf{x},y) \log \varphi(\mathbf{x},y) \right]}_{:= B} \\
& \underbrace{+\alpha \mathbb{E}_{P_I(\mathbf{x},y)} \left[ \log \sigma (\log \varphi(\mathbf{x},y) ) \right]}_{:= C} \\
& \underbrace{-\alpha  \mathbb{E}_{P(\mathbf{x})} \mathbb{E}_{P_I(y)} \left[ \log \left( 1 - \sigma (\log \varphi(\mathbf{x},y) ) \right)\right]}_{:= D}.\\
\end{aligned}
\end{equation}
However, the complex objective function Eq.~(\ref{eq:SA2}) causes the high cost of calculating the gradients for optimization. To simplify the above function, we first obtain the following lower bound,
\begin{equation}
\begin{aligned}
\label{eq:AC}
A + C= & (1 - \alpha) \mathbb{E}_{P_I(\mathbf{x},y)} \left[ \log \varphi(\mathbf{x},y) \right] \\
& + \alpha \mathbb{E}_{P_I(\mathbf{x},y)} \left[ \log \frac{\varphi(\mathbf{x},y)}{\varphi(\mathbf{x},y) + 1} \right] \\
= & \mathbb{E}_{P_I(\mathbf{x},y)} \left[ \log \varphi(\mathbf{x},y) \right] \\
& - \alpha \mathbb{E}_{P_I(\mathbf{x},y)} \left[ \log \left( \varphi(\mathbf{x},y) + 1\right) \right] \\
\geq & \mathbb{E}_{P_I(\mathbf{x},y)} \left[ \log \varphi(\mathbf{x},y) \right] - \alpha \log 2,
\end{aligned}
\end{equation}
where the first equality is due to the property of the sigma function,
\begin{equation}
\begin{aligned}
\label{eq:sigma1}
\sigma( \log s) = \frac{s}{s + 1}, s \geq 0,
\end{aligned}
\end{equation}
and the first inequality holds since $\varphi(\mathbf{x},y) \in (0,1]$. Also, we have the following lower bound,
\begin{equation}
\begin{aligned}
\label{eq:BD}
 B + D \geq & - (1 - \alpha) \beta \mathbb{E}_{P(\mathbf{x})} \left[ \sum_{y \in [K]} \varphi(\mathbf{x},y) \log \varphi(\mathbf{x},y) \right] \\
& + \alpha  \mathbb{E}_{P(\mathbf{x})} \mathbb{E}_{P_I(y)} \left[ \log \left( \varphi(\mathbf{x},y) + 1 \right)\right] \\
 \geq & \mathbb{E}_{P(\mathbf{x})} \left[ \sum_{y \in [K]} \alpha P_I(y) \log \varphi(\mathbf{x},y) \right]\\
 & -  \mathbb{E}_{P(\mathbf{x})} \left[ \sum_{y \in [K]}  (1 - \alpha) \beta \varphi(\mathbf{x},y)  \log \varphi(\mathbf{x},y) \right],
\end{aligned}
\end{equation}
where the first inequality holds owing to Eq.~(\ref{eq:sigma1}), and the second inequality holds since $\log(x)$ is a monotonically increasing function. To obtain a compact result, we assume $\beta = \alpha / (1 - \alpha)$ without loss of generality. Substituting Eq.~(\ref{eq:AC}) and Eq.~(\ref{eq:BD}) into Eq.~(\ref{eq:SA2}), we derive the objective function for the SA method:
\begin{equation}
\begin{aligned}
\label{eq:SA}
%\mathcal{L}_{SA}(\theta) = & \mathbb{E}_{P_I(\mathbf{x},y)} \left[ \log \varphi(\mathbf{x},y) \right] \\
%& + \alpha \mathbb{E}_{P(\mathbf{x})} \left[ \sum_{y  \in [K]} \left(  P_I(y) - \varphi(\mathbf{x},y) \right) \log \varphi(\mathbf{x},y) \right],
\mathcal{L}_{SA}(\theta) = & \mathbb{E}_{P_I(\mathbf{x},y)} \left[ \log \varphi(\mathbf{x},y) \right] + \alpha \mathbb{E}_{P(\mathbf{x})} \mathcal{R}(\mathbf{x}),\\
\end{aligned}
\end{equation}
where
\begin{equation}
\mathcal{R}(\mathbf{x}) = \sum_{y  \in [K]} \left(  P_I(y) - \varphi(\mathbf{x},y) \right) \log \varphi(\mathbf{x},y)
\end{equation}
and $P_I(y)$ is an ID class probability, which can be estimated by exploring ID samples in experiments. From the derived result in Eq.~(\ref{eq:SA}), it becomes evident that the adaptive supervision information for any given OOD sample relative to class $y$ is $P_I(y) - \varphi(\mathbf{x},y)$. This is achieved through the seamless amalgamation of MSMI and MBCE components. The complete algorithmic procedure is succinctly delineated in Algorithm~\ref{alg:SA}.

\section{Experimental Setup} \label{sec:setup}
In this section, we elucidate the network architectures employed by various OOD detection methods. Additionally, we delineate both the ID and OOD datasets selected for experimental validation, as well as the metrics employed for performance evaluation.

\subsection{Network Architectures}
To demonstrate the wide-ranging applicability of our proposed SA method, we integrate it into four advanced neural networks: ResNet18~\cite{RES:16}, VGG19~\cite{VGG:15}, MobileNetV2~\cite{MN:18}, and EfficientNet~\cite{EN:19}. These networks are implemented using the PyTorch framework and are trained on a single GPU. Each network undergoes training for $200$ epochs with mini-batches containing $128$ examples, leveraging the stochastic gradient descent (SGD) optimization. In line with the configuration used in the foundational Mixup method~\cite{MIXUP:18}, we initiate the learning rates at $0.1$, subsequently reducing them by a factor of $10$ at the $100$ and $150$ epochs for all models. Additionally, we enforce a weight decay of $10^{-4}$, while abstaining from employing dropout in our experiments.

\subsection{Datasets}
To validate the robustness and effectiveness of our proposed SA methodology, we train neural networks on two ID datasets: CIFAR10 and CIFAR100~\cite{CIFAR10:09}. Both datasets feature $60,000$ RGB images of dimensions $32 \times 32 \times 3$, partitioned into $50,000$ training and $10,000$ test samples. The CIFAR10 and CIFAR100 datasets have 10 and 100 classes, respectively. For ID training, we utilize standard data augmentation techniques, including random crops with a zero-padding length of $4$ and random horizontal flips.

To measure the OOD detection performance, we employ test samples from the training ID datasets as ID evaluation sets. For OOD evaluation, we draw samples from nine real-world image datasets as well as two synthetic noise datasets: SVHN~\cite{SVHN:11}, iSUN~\cite{ISUN:15}, LSUN~\cite{LSUN:15}, TinyImageNet~\cite{IMAGENET:09}, CelebA~\cite{CLA:15}, VisDA~\cite{VisDA:17}, Caltech256~\cite{CAL:06}, PASC~\cite{PASC:17}, COCO~\cite{COCO:14}, Gaussian, and Uniform. For a training dataset containing $M$ in-distribution samples, and according to our defined mixture distribution $P(\mathbf{x})$, the number of OOD samples required for training is computed as $ N = M \times \epsilon / (1 - \epsilon)$. Any remaining OOD samples are reserved for testing OOD detection capabilities. For datasets not used in training, all their samples serve as test OOD sets. To maintain consistency, all OOD images are cropped to $32 \times 32 \times 3$, matching the dimensions of the ID samples.

Further details about the OOD datasets can be found in \tablename~\ref{tb:ddd}. Notably, the iSUN and Caltech256 datasets, which lack official train-test splits, are used entirely for OOD testing. For the Gaussian and Uniform datasets, we synthesize samples of dimensions $32 \times 32 \times 3$, drawing independent and identically distributed values from standard normal and uniform distributions, respectively. To ensure fair comparison, datasets like CelebA and VisDA, which contain a large number of samples, are downsampled to $10,000$ randomly selected instances. For the COCO dataset, we utilize the validation set as the OOD sample source. The training set from VisDA is employed, given the unavailability of its test set.

\begin{table*} \tiny
  \renewcommand{\arraystretch}{1.3}
  \centering
  \caption{Description of OOD datasets. Symbol $\#$ indicates `the number of'.}
   \label{tb:ddd}
\begin{tabular}{cccccccccccc}
\toprule
Dataset & SVHN & iSUN & LSUN & TinyImageNet & CelebA & Caltech256 & VisDA & PASC & COCO & Gaussian & Uniform \\ \midrule \midrule
\# samples & 26,032 & 10,000 & 10,000 & 10,000 & \begin{tabular}[c]{@{}c@{}}10,000\\ (Randomly Selected)\end{tabular} & 30,607 & \begin{tabular}[c]{@{}c@{}}10,000\\ (Randomly Selected)\end{tabular} & 9991 & 5000 & 10,000 & 10,000 \\ \midrule
Source & Test Set & Entire Set & Test Set & Test Set & Test Set & Entire Set & \begin{tabular}[c]{@{}c@{}}Training Set\\ (2017 classification track)\end{tabular} & Test Set & \begin{tabular}[c]{@{}c@{}}Validation Set\\ (2017 Object Detection Task)\end{tabular} & $\mathcal{N}(0,1)$ & $\mathcal{U}(0,1)$ \\ \bottomrule
\end{tabular}
\end{table*}

\subsection{Metrics}
To evaluate the proposed SA method on various fronts, including efficiency, ID sample classification, and OOD detection performance, we use throughput~\cite{TP:21}, accuracy (ACC), and the area under the precision-recall curve (AUROC)~\cite{AUROC:06} as evaluation metrics. Throughput, quantified as the number of samples processed per second, serves as an indicator of computational efficiency during the optimization of objective functions. Accuracy (ACC) offers a measure of the performance on classifying ID samples. AUROC, a threshold-independent metric, gauges the likelihood that an ID sample receives higher predictive confidence than an OOD sample, particularly when dealing with imbalanced data. As AUROC is derived from the Receiver Operating Characteristic (ROC) curve, it also encapsulates the trade-off between the true positive rate (TPR) and false positive rate (FPR) across different decision thresholds. To further elucidate the significance of AUROC, we present the ROC curve for the ResNet18 model trained on CIFAR100 against OOD samples from the TinyImageNet dataset. This curve is depicted in \figurename~\ref{fig:curve}. A larger AUROC value signifies superior performance in detecting OOD samples. In our experiments, the SA method outperforms the random method, which has an AUROC score of $1/2$, indicating its robustness in OOD detection. For a comprehensive comparison of all methods tested on various OOD datasets, we adhere to the baseline setup~\cite{BL:17}, unless otherwise specified, using the maximum probability generated by the softmax output as the OOD score for each sample.

\section{Comparison Experiments} \label{sec:comRes}
We evaluate the performance of the proposed SA method\footnote{The source codes are available at: \url{https://github.com/Lawliet-zzl/SA}.} across a range of benchmarks, comparing it against the established techniques in various sub-fields of machine learning, including OOD detection, OOD generalization, domain adaptation, and generalization improvement. For the purpose of OOD detection, we consider methods such as the baseline~\cite{BL:17}, OE~\cite{OE:19}, ACET~\cite{ADB:19}, GODIN~\cite{DCC:20}, EBD~\cite{EB:20}, MOS~\cite{MOS:21}, and FIG~\cite{Z22ood} as benchmarks. In the realm of OOD generalization, our comparisons include with methods such as RSC~\cite{RSC:20}, IRM~\cite{IRM:19}, KerHRM~\cite{KHRM:21}, and GAM~\cite{GNA:23}. Domain adaptation methods in our benchmark suite encompass DANN~\cite{DANN:16} and DCORAL~\cite{CORAL:16}. Lastly, for evaluating generalization improvement, we consider methods including Gaussian noise (GN)~\cite{GN:17}, MIXUP~\cite{MIXUP:18}, LS~\cite{CP:17}, and Tf-KD~\cite{KD:20}. Through this comprehensive benchmarking, we aim to demonstrate the versatility and effectiveness of the SA method in pursuing a better balance between high-quality ID classification and OOD detection.

In our experiments, we adhere to the configurations proposed in the original papers for all the comparison methods to ensure a fair and meaningful evaluation. We present the results using four state-of-the-art neural network architectures: ResNet18, VGG19, MobileNetV2, and EfficientNet, trained on two widely-used ID datasets, namely CIFAR10 and CIFAR100. For methods that incorporate OOD samples in the training phase, TinyImageNet is utilized as the default OOD dataset, unless otherwise specified. We set the hyperparameter $\epsilon = 0.05$ to standardize comparisons. For domain adaptation techniques, such as CORAL and DANN, TinyImageNet is treated as the target domain data. For our proposed SA method, the parameter $\alpha$ is set to $0.2$, which was found to offer a balanced trade-off between high ID classification accuracy and OOD detection capabilities. The performance metrics employed in our experiments include ACC on the test set of the corresponding ID dataset and the average AUROC across eleven different OOD datasets. These OOD datasets include SVHN, iSUN, LSUN, TinyImageNet, CelebA, Caltech256, VisDA, PASC, COCO, Gaussian, and Uniform. The results of our experiments are detailed in \tablename~\ref{tb:ce}. To provide a reliable evaluation, all ACC and AUROC values reported are averaged over five independent trials.

\begin{table*}[]
  \renewcommand{\arraystretch}{1.3}
  \renewcommand\tabcolsep{2.0pt}
  \centering
  \caption{OOD detection and classification performance. All reported values are averaged over five trials. Each AUROC value indicates the average score across the eleven OOD datasets. Boldface values indicate the relatively better performance. Symbol $-$, $\uparrow$, and $\downarrow$ indicate the performance is equal, superior and inferior to that of the baseline method.}
  \label{tb:ce} %\vspace{-0.6em}
\begin{tabular}{ccccccccccc}  \toprule
\multirow{2}{*}{In-dist}                       & \multirow{2}{*}{Category}               & \multirow{2}{*}{Method} & \multicolumn{2}{c}{ResNet18} & \multicolumn{2}{c}{VGG19} & \multicolumn{2}{c}{MobileNetV2} & \multicolumn{2}{c}{EffificientNet} \\ \cline{4-11}
                                               &                                         &                         & AUROC         & ACC          & AUROC        & ACC        & AUROC           & ACC           & AUROC            & ACC             \\  \midrule\midrule
\multirow{18}{*}{CIFAR10}                      & \multirow{7}{*}{OOD Detection}
& Baseline     & 91.4$_{\pm 0.0} - $             & 95.0$_{\pm 0.0} -$                   & 89.3$_{\pm 0.0} - $              & 93.5$_{\pm 0.0} - $                 & 87.9$_{\pm 0.0} - $             & 91.3$_{\pm 0.0} - $                 & 91.3$_{\pm 0.0} - $              & 90.6$_{\pm 0.0} - $            \\
& & OE           & 95.0$_{\pm 0.1} \uparrow$ & 94.9$_{\pm 0.1} \downarrow$ & 92.9$_{\pm 0.1} \uparrow $  & 92.7$_{\pm 0.2} \downarrow$ & 93.9$_{\pm 0.3} \uparrow$ & 90.7$_{\pm 1.1} \downarrow$ & 92.5$_{\pm 0.0} \uparrow$ & 90.2$_{\pm 0.3} \downarrow$ \\
& & ACET & 96.5$_{\pm 0.1} \uparrow$  & 94.6$_{\pm 0.2} \downarrow$  & 95.2$_{\pm 0.5} \uparrow$ & 92.7$_{\pm 0.1} \downarrow$ & 94.0$_{\pm 0.2} \uparrow$ & 90.7$_{\pm 0.1} \downarrow$ & 93.7$_{\pm 0.0} \uparrow$ & 90.0$_{\pm 0.1} \downarrow$  \\
& & GODIN       & 96.2$_{\pm 0.1} \uparrow$ & 93.2$_{\pm 0.3} \downarrow$ & 95.6$_{\pm 0.1} \uparrow $  & 92.3$_{\pm 0.1} \downarrow$ & 95.3$_{\pm 0.4} \uparrow$ & 91.0$_{\pm 0.1} \downarrow$ & 94.9$_{\pm 0.1} \uparrow$ & 89.4$_{\pm 0.0} \downarrow$ \\
& & EBD         & 93.7$_{\pm 0.0} \uparrow$ & 95.0$_{\pm 0.0} -$                  & 92.6$_{\pm 0.0} \uparrow $ & 93.5$_{\pm 0.0} - $                   & 89.7$_{\pm 0.0} \uparrow$ & 91.3$_{\pm 0.0} -$                  & 93.2$_{\pm 0.0} \uparrow$ & 90.6$_{\pm 0.0} - $ \\
& & MOS        & 92.3$_{\pm 0.4} \uparrow$ & 94.5$_{\pm 0.1} \downarrow$ & 93.4$_{\pm 0.0} \uparrow $ & 92.1$_{\pm 0.2} \downarrow$   & 90.8$_{\pm 0.1} \uparrow$ & 90.5$_{\pm 0.6} \downarrow$ & 93.6$_{\pm 0.0} \uparrow$ & 88.4$_{\pm 0.1} \downarrow$ \\
& & FIG           & 97.1$_{\pm 0.3} \uparrow$ & 95.0$_{\pm 0.1} - $                 & 95.9$_{\pm 0.1} \uparrow $ & 93.2$_{\pm 0.0} \downarrow$   & 95.2$_{\pm 0.1} \uparrow$ & 91.0$_{\pm 0.1} \downarrow$ & 95.1$_{\pm 0.2} \uparrow$ & 90.3$_{\pm 0.1} \downarrow$ \\  \cline{2-11}
& \multirow{6}{*}{\begin{tabular}[c]{@{}c@{}}OOD Generalization \\and Domain Adaption\end{tabular}}
& RSC          & 92.1$_{\pm 0.1} \uparrow$ & 95.0$_{\pm 0.9} -$                       & 92.3$_{\pm 0.5} \uparrow    $ & 93.2$_{\pm 0.1} \downarrow $ & 89.1$_{\pm 0.3} \uparrow$ & 91.0$_{\pm 0.1} \downarrow$ & 93.5$_{\pm 0.0} \uparrow$ & 90.6$_{\pm 0.2} -$ \\
& & IRM           & 93.2$_{\pm 0.2} \uparrow$ & 94.8$_{\pm 0.1} \downarrow $ & 93.1$_{\pm 0.3} \uparrow        $ & 92.9$_{\pm 0.3} \downarrow $ & 92.8$_{\pm 0.3} \uparrow$ & 90.7$_{\pm 0.6} \downarrow$ & 93.3$_{\pm 0.1} \uparrow$ & 90.5$_{\pm 0.2} \downarrow$ \\
& & KerHRM           & 93.8$_{\pm 0.1} \uparrow$ & 94.9$_{\pm 0.2} \downarrow$ & 92.4$_{\pm 0.3} \uparrow$ & 93.4$_{\pm 0.1} \downarrow$ & 91.7$_{\pm 0.2} \uparrow$ & 91.4$_{\pm 0.2} \uparrow$ & 93.9$_{\pm 0.1} \uparrow$ & 90.4$_{\pm 0.1} \downarrow$ \\
& & GAM           & 94.2$_{\pm 0.0} \uparrow$ & 95.0$_{\pm 0.2} -$ & 93.5$_{\pm 0.1} \uparrow$ & 93.3$_{\pm 0.3} \downarrow$ & 92.1$_{\pm 0.1} \uparrow$ & 91.3$_{\pm 0.2}-$ & 93.8$_{\pm 0.5} \uparrow$ & 90.6$_{\pm 0.1}-$ \\
& & DANN       & 87.1$_{\pm 0.1} \downarrow$ & 94.6$_{\pm 0.1} \downarrow$ & 82.1$_{\pm 0.1} \downarrow$ & 92.1$_{\pm 0.3} \downarrow $ & 85.2$_{\pm 0.0} \downarrow$ & 90.4$_{\pm 0.7} \downarrow$ & 84.4$_{\pm 0.2} \downarrow$ & 90.0$_{\pm 0.1} \downarrow$\\
& &DCORAL        & 88.5$_{\pm 0.3} \downarrow$ & 93.7$_{\pm 0.1} \downarrow$ & 80.3$_{\pm 0.2} \downarrow$ & 93.2$_{\pm 0.2} \downarrow $ & 86.3$_{\pm 0.5} \downarrow$ & 91.3$_{\pm 0.1} - $     & 86.4$_{\pm 0.4} \downarrow$ & 89.9$_{\pm 0.2} \downarrow$ \\  \cline{2-11}
& \multirow{4}{*}{Generation Improvement}
& GN                 & 90.5$_{\pm 0.2} \downarrow$ & 94.7$_{\pm 0.2} \downarrow$ & 85.2$_{\pm 0.1} \downarrow$ & 92.9$_{\pm 0.1} \downarrow $ & 86.7$_{\pm 0.2} \downarrow$ & 90.8$_{\pm 0.2} \downarrow$ & 87.8$_{\pm 0.5} \downarrow $ & 89.9$_{\pm 0.1} \downarrow$ \\
& & MIXUP       & 91.7$_{\pm 0.7} \uparrow$ & \textbf{95.8}$_{\pm 0.3} \uparrow$ & 92.9$_{\pm 0.1} \uparrow  $ & \textbf{94.3}$_{\pm 0.2} \uparrow$ & 89.1$_{\pm 0.1} \uparrow$ & 91.3$_{\pm 0.2} - $               & 87.2$_{\pm 0.1} \downarrow$ & 88.9$_{\pm 0.1} \downarrow$ \\
& & LS              & 86.7$_{\pm 0.1} \downarrow$ & 95.1$_{\pm 0.1} \uparrow$ & 76.8$_{\pm 0.4} \downarrow     $ & 93.3$_{\pm 0.4} \downarrow $ & 86.3$_{\pm 0.0} \downarrow$ & 91.2$_{\pm 0.1} \downarrow$ & 86.5$_{\pm 0.8} \downarrow$ & 90.6$_{\pm 0.1} - $ \\
& & Tf-KD         & 87.9$_{\pm 0.0} \downarrow$ & 95.2$_{\pm 0.1} \uparrow$ & 77.8$_{\pm 0.1} \downarrow     $ & 93.2$_{\pm 0.1} \downarrow $ & 86.5$_{\pm 0.1} \downarrow$ & 91.0$_{\pm 0.0} \downarrow$ & 82.1$_{\pm 0.0} \downarrow$ & 89.3$_{\pm 0.1} \downarrow$ \\ \cline{2-11}
& Ours  & SA    & \textbf{98.9}$_{\pm 0.1} \uparrow$ & 95.1$_{\pm 0.1} \uparrow$ & \textbf{96.4}$_{\pm 0.2} \uparrow $ & 93.7$_{\pm 0.1} \uparrow $ & \textbf{97.6}$_{\pm 0.1} \uparrow$ & \textbf{91.6}$_{\pm 0.3} \uparrow$ & \textbf{97.0}$_{\pm 0.2} \uparrow$ & \textbf{90.7}$_{\pm 0.1} \uparrow$ \\  \midrule\midrule
\multicolumn{1}{l}{\multirow{18}{*}{CIFAR100}} & \multirow{7}{*}{OOD Detection}
&   Baseline & 80.1$_{\pm 0.0} - $             & 77.7$_{\pm 0.0} - $ & 72.0$_{\pm 0.0} - $ & 71.4$_{\pm 0.0} - $ & 72.6$_{\pm 0.0} - $ & 71.2$_{\pm 0.0} - $ & 74.2$_{\pm 0.0} - $ & 69.1$_{\pm 0.0}-$ \\
& & OE        & 89.0$_{\pm 0.5} \uparrow$ & 77.6$_{\pm 0.3} \downarrow$ & 86.7$_{\pm 0.8} \uparrow$ & 72.0$_{\pm 0.2} \uparrow$ & 88.3$_{\pm 0.8} \uparrow$ & 70.8$_{\pm 1.0} \downarrow$ & 87.2$_{\pm 0.9} \uparrow$ & 68.7$_{\pm 1.5} \downarrow$ \\
& & ACET & 92.6$_{\pm 0.1} \uparrow $  & 77.5$_{\pm 0.2} \downarrow $ & 87.8$_{\pm 0.0} \uparrow $ & 71.9$_{\pm 0.0} \uparrow $ & 89.4$_{\pm 0.1} \uparrow $ & 70.6$_{\pm 0.0} \downarrow $ & 89.1$_{\pm 0.1} \uparrow $ & 68.6$_{\pm 0.3} \downarrow $  \\
& & GODIN      & 93.9$_{\pm 0.5} \uparrow$ & 76.3$_{\pm 0.2} \downarrow$ & 89.6$_{\pm 0.3} \uparrow$ & 71.6$_{\pm 0.1} \uparrow$ & 90.8$_{\pm 0.6} \uparrow$ & 70.4$_{\pm 0.8} \downarrow$ & 92.7$_{\pm 0.6} \uparrow$ & 68.8$_{\pm 1.2} \downarrow$ \\
& & EBD      & 89.7$_{\pm 0.0} \uparrow$ & 77.7$_{\pm 0.0} -$                  & 80.6$_{\pm 0.0} \uparrow$ & 71.4$_{\pm 0.0} - $             & 89.9$_{\pm 0.0} \uparrow$ & 71.2$_{\pm 0.0} - $              & 86.5$_{\pm 0.0}\uparrow$ & 69.1$_{\pm 0.0} - $ \\
& & MOS     & 86.7$_{\pm 0.2} \uparrow$ & 76.1$_{\pm 0.5}\downarrow$ & 85.6$_{\pm 0.6} \uparrow$ & 71.0$_{\pm 0.5} \downarrow$ & 88.9$_{\pm 0.5} \uparrow$ & 70.8$_{\pm 1.2} \downarrow$ & 90.1$_{\pm 0.6} \uparrow$ & 68.1$_{\pm 0.4} \uparrow$ \\
& & FIG       & 94.3$_{\pm 0.6} \uparrow$ & 77.5$_{\pm 0.3}\downarrow$ & 90.7$_{\pm 0.7}\uparrow $ & 71.2$_{\pm 0.7} \downarrow$ & 91.1$_{\pm 0.2} \uparrow$ & 71.0$_{\pm 0.4} \downarrow$ & 92.0$_{\pm 0.3} \uparrow$ & 68.5$_{\pm 0.5} \uparrow$ \\ \cline{2-11}
& \multirow{6}{*}{\begin{tabular}[c]{@{}c@{}}OOD Generalization \\and Domain Adaption\end{tabular}}
& RSC       & 85.1$_{\pm 0.4} \uparrow$ & 78.2$_{\pm 0.9}\uparrow$         & 83.4$_{\pm 0.6} \uparrow$ & 71.9$_{\pm 0.7} \uparrow$ & 85.5$_{\pm 0.4} \uparrow$ & 71.8$_{\pm 0.7} \uparrow$ & 85.8$_{\pm 0.1} \uparrow$ & 69.4$_{\pm 0.2} \uparrow$ \\
& & IRM       & 84.4$_{\pm 1.1} \uparrow$ & 76.5$_{\pm 0.4}\downarrow$     & 87.3$_{\pm 0.8} \uparrow$ & 71.1$_{\pm 0.5} \downarrow$ & 87.6$_{\pm 0.1} \uparrow$ & 70.9$_{\pm 0.2} \downarrow$ & 83.8$_{\pm 0.1} \uparrow$ & 68.6$_{\pm 0.6} \uparrow$ \\
& & KerHRM           & 87.6$_{\pm 0.3} \uparrow$ & 78.3$_{\pm 0.2} \uparrow$ & 85.7$_{\pm 0.3} \uparrow$ & 72.0$_{\pm 0.2} \uparrow$ & 88.3$_{\pm 0.4} \uparrow$ & 72.0$_{\pm 0.1} \uparrow$ & 86.5$_{\pm 0.3} \uparrow$ & 69.3$_{\pm 0.1} \uparrow$ \\
& & GAM           & 86.2$_{\pm 0.1} \uparrow$ & 78.4$_{\pm 0.3} \uparrow$ & 87.5$_{\pm 0.1} \uparrow$ & 72.2$_{\pm 0.2} \uparrow$ & 89.0$_{\pm 0.2} \uparrow$ & 71.9$_{\pm 0.5} \uparrow$ & 88.0$_{\pm 0.1} \uparrow$ & 69.3$_{\pm 0.1} \uparrow$ \\
& & DANN    & 77.9$_{\pm 0.4} \downarrow$ & 77.9$_{\pm 1.1} \uparrow$    & 67.1$_{\pm 1.8} \downarrow$ & 71.6$_{\pm 1.5} \uparrow$ & 69.3$_{\pm 0.5} \downarrow$ & 71.5$_{\pm 0.9} \uparrow$ & 70.9$_{\pm 1.0} \downarrow $ & 69.1$_{\pm 0.6} - $ \\
& & DCORAL     & 78.2$_{\pm 0.5} \downarrow$ & 77.1$_{\pm 0.3}\downarrow$ & 68.2$_{\pm 0.9} \downarrow$ & 70.9$_{\pm 1.7} \downarrow$ & 70.9$_{\pm 1.1} \downarrow$ & 70.6$_{\pm 0.2} \downarrow$ & 71.2$_{\pm 0.2} \downarrow$ & 68.9$_{\pm 0.1} \uparrow$ \\
\cline{2-11}
& \multirow{4}{*}{Generation Improvement}
& GN             & 81.5$_{\pm 1.2} \uparrow$ & 76.5$_{\pm 0.3}\downarrow$ & 74.0$_{\pm 0.1} \uparrow$ & 70.4$_{\pm 0.1} \uparrow$ & 72.7$_{\pm 1.4} \uparrow$ & 69.1$_{\pm 0.5} \downarrow$ & 77.9$_{\pm 1.3} \uparrow$ & 67.7$_{\pm 0.2} \uparrow$ \\
& & MIXUP   & 82.0$_{\pm 0.8} \uparrow$ & 78.2$_{\pm 0.3} \uparrow$ & 77.5$_{\pm 1.2} \uparrow$ & \textbf{72.8}$_{\pm 0.2} \uparrow$ & 77.5$_{\pm 0.8} \uparrow$ & 68.6$_{\pm 0.0} \downarrow$ & 74.8$_{\pm 1.5} \uparrow$ & 66.2$_{\pm 0.1} \uparrow$ \\
& & LS          & 83.2$_{\pm 0.6} \uparrow$ & 78.8$_{\pm 0.5} \uparrow$ & 75.8$_{\pm 0.4} \uparrow$ & \textbf{72.8}$_{\pm 0.5} \uparrow$ & 78.6$_{\pm 0.4} \uparrow$ & 71.0$_{\pm 0.5} \downarrow$ & 80.4$_{\pm 0.2} \uparrow$ & 69.0$_{\pm 0.1} \uparrow$ \\
& & Tf-KD     & 80.1$_{\pm 1.3} \uparrow$ & \textbf{79.0} $_{\pm 0.4}\uparrow$ & 75.4$_{\pm 0.3} \uparrow$ & 72.4$_{\pm 0.2} \uparrow$ & 68.9$_{\pm 0.8} \downarrow$ & 71.6$_{\pm 0.3} \uparrow$ & 79.3$_{\pm 0.7} \uparrow$ & 69.1$_{\pm 0.5}  -$ \\ \cline{2-11}
& Ours & SA & \textbf{96.3}$_{\pm 0.6}  \uparrow$ & 78.6$_{\pm 0.4}\uparrow$ & \textbf{94.4}$_{\pm 1.2} \uparrow$ & 72.7$_{\pm 0.1} \uparrow$ & \textbf{92.7}$_{\pm 0.2} \uparrow$ & \textbf{72.2}$_{\pm 0.3} \uparrow$ & \textbf{94.1}$_{\pm 0.5} \uparrow$ & \textbf{69.6}$_{\pm 0.1} \uparrow$ \\ \bottomrule
\end{tabular} %\vspace{-0.6em}
\end{table*}

\subsection{Comparison for OOD Detection}~\label{sec:comOOD}
Our experimental results reveal noteworthy observations. All OOD detection methods under consideration have significantly outperformed the baseline method in terms of OOD detection capabilities. Among these, our proposed SA method stands out by delivering the best performance, registering a marked improvement in OOD detection scores ranging from $0.83\%$ to $31.11 \%$ over the state-of-the-art alternatives. The performance of SA is attributed to the ability to adaptively generate supervision information for OOD samples, thereby facilitating a more nuanced and accurate characterization of these samples. This enables the neural network to better generalize from one OOD dataset to potentially unseen OOD samples, effectively improving the network sensitivity to OOD samples. Interestingly, while most methods showed a trade-off between improving OOD detection and maintaining high classification accuracy on ID samples, SA has managed to excel in both aspects. Our results indicate an improvement in ID classification accuracy ranging from $0.11 \%$ to $1.96 \%$, even as SA boosts OOD detection. This can be attributed to the design philosophy of SA, which avoids the pitfalls of manual labeling for OOD samples that could otherwise distort the data distribution and compromise the focus of networks on the classification task. SA intelligently adapts the supervision information for OOD samples in a way that minimizes their impact on ID sample classification, thereby allowing for a more refined separation of ID classes and ultimately enhancing the generalization capabilities. In summary, our SA method successfully negotiates the balance between two conflicting objectives: robust OOD detection and high-quality ID classification. This demonstrates that SA offers a more effective and nuanced trade-off between these two goals compared to existing state-of-the-art OOD detection methods.

\begin{figure*}
  \centering
  \includegraphics[width=0.95\textwidth]{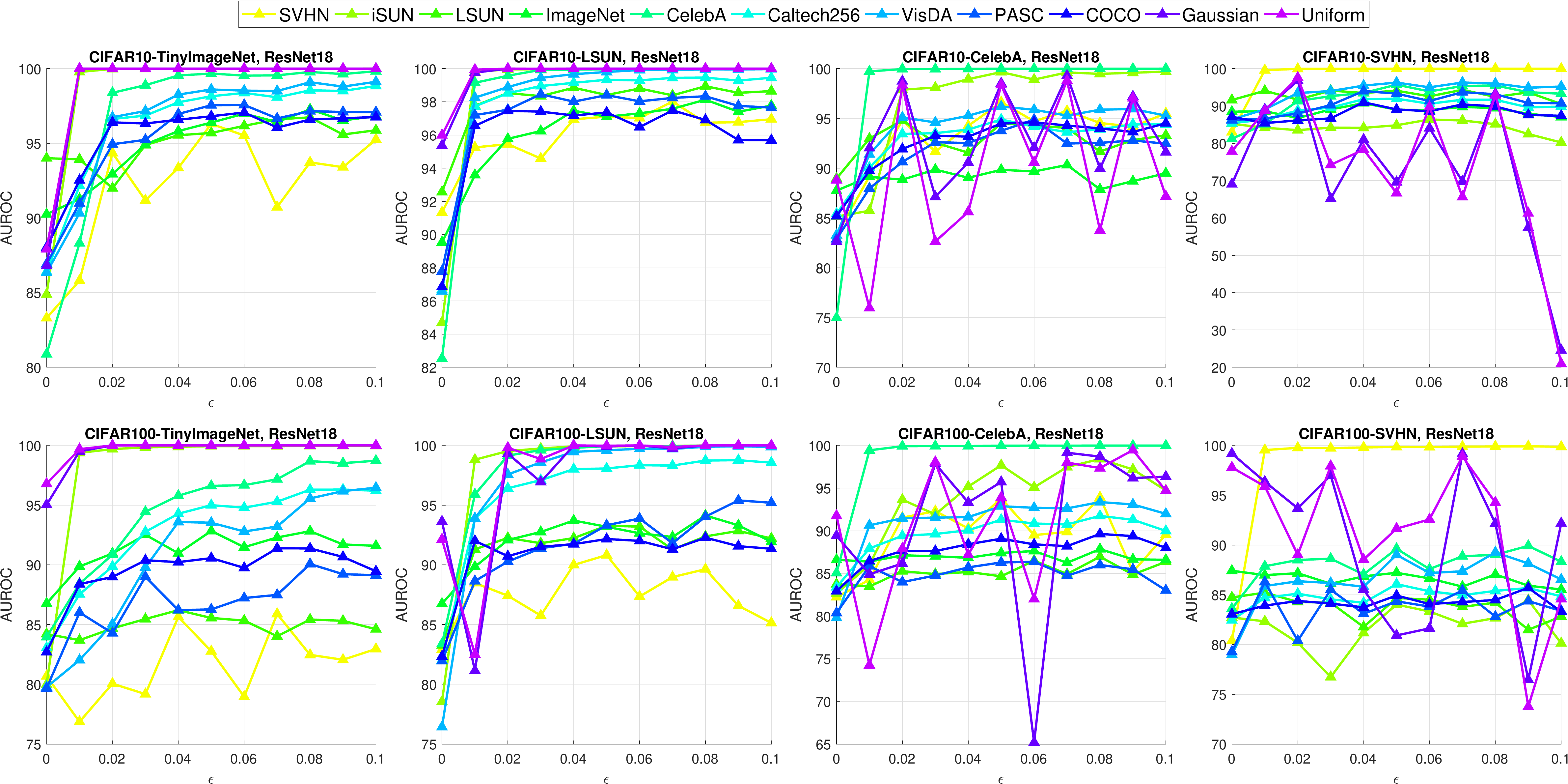}
  \caption{Effect of component parameter $\epsilon$ when  combination parameter $\alpha = 0.2$. Each title includes the corresponding training ID dataset, training OOD dataset, and network architecture. Eleven different OOD datasets are used to test the OOD detection performance, and each broken line corresponds to a test OOD dataset. Each point indicates an AUROC score on a test OOD dataset, and larger values are better.}
  \label{fig:eps}
\end{figure*}

\subsection{Comparison for OOD Generalization and Domain Adaptation}~\label{sec:comDG}
The OOD generalization methods, including RSC, IRM, KerHRM, and GAM, demonstrate better OOD detection performance compared to the baseline. In contrast, domain adaptation methods like CORAL and DANN exhibit a decline in the OOD detection efficacy. The former group enhances OOD detection by focusing on invariant features and identifying spurious attributes that prevent high-confidence misclassifications of OOD samples. The latter group, however, aligns the distributions of ID and OOD samples, inadvertently making the network more likely to incorrectly assign high-confidence predictions to OOD samples. When compared with both OOD generalization and domain adaptation methods, our SA method achieves a superior balance, showing improvements in ID classification accuracy ranging from $0.11 \%$ to $ 2.75 \%$ and in OOD detection performance from $3.54 \%$ to $ 40.69 \%$. This is partly because SA effectively manages OOD samples. It not only discourages the network from making high-confidence errors for these samples but also uses their unique characteristics to boost the ability to generalize on ID classes. By creating adaptive supervision information, SA improves the distinguishability of ID classes, thereby enhancing both ID and OOD performance.

\subsection{Comparison for Generalization Improvement}~\label{sec:comGI}
We observe that all generalization improvement methods fall short in OOD sample detection compared to the baseline. This occurs because these methods tend to reduce the prediction confidence for ID samples without adequately constraining OOD samples, thereby blurring the distinction between the two. However, the SA method is different: while it also encourages smooth output probabilities for ID samples, it explicitly aims to lower the confidence predictions for OOD samples. This is achieved by leveraging the design principle of the MBCE component to distinguish between ID and OOD samples effectively. As a result, SA realizes a substantial improvement in OOD detection performance, ranging from $3.70 \%$ to $34.54 \%$ compared to generalization improvement methods. Regarding classification accuracy, SA is on par with other generalization methods and outperforms the baseline by $0.22 \%$ and $1.26 \%$ on CIFAR10 and CIFAR100, respectively. This can be attributed mainly to two factors: (1) the adaptiveness of the supervision information ensures that the presence of OOD samples does not interfere with the learning process for ID samples, and (2) the use of an OOD dataset to better differentiate ID samples with varying labels contributes to further improvement.

\section{Quantitative and Qualitative Analyses}~\label{sec:QA}
In this section, we explore the influence of various parameters and the datasets used for OOD training. We also conduct a comprehensive ablation study, evaluate model efficiency, and present visual insights into our findings.

\subsection{Effect of Parameters}~\label{sec:eop}
We investigate the impact of the component parameter $\epsilon$ and the combination parameter $\alpha$ in the SA method on performance metrics, specifically focusing on ACC and AUROC.

\subsubsection{Effect of Component Parameters}
We examine the influence of the component parameter $\epsilon$ in SA by analyzing $11$ evenly-spaced values ranging from $0$ to $0.1$. We set $\alpha$ to a fixed value of $0.2$ and use ResNet18 as the network architecture. This is visualized in \figurename~\ref{fig:eps}. Our findings indicate that a larger $\epsilon$ generally enhances OOD detection performance, though the benefit tends to plateau when the parameter becomes overly large. Notably, even a modest $\epsilon$, such as 0.05, can yield high AUROC scores. This suggests that even a small-scale training dataset of OOD samples can suffice to assure strong OOD detection performance. This efficacy is primarily attributed to the ability of SA to make the most of adaptive supervision information, thereby extracting valuable insights from a limited pool of training OOD samples.

\begin{figure*}
  \centering
  \includegraphics[width=0.90\textwidth]{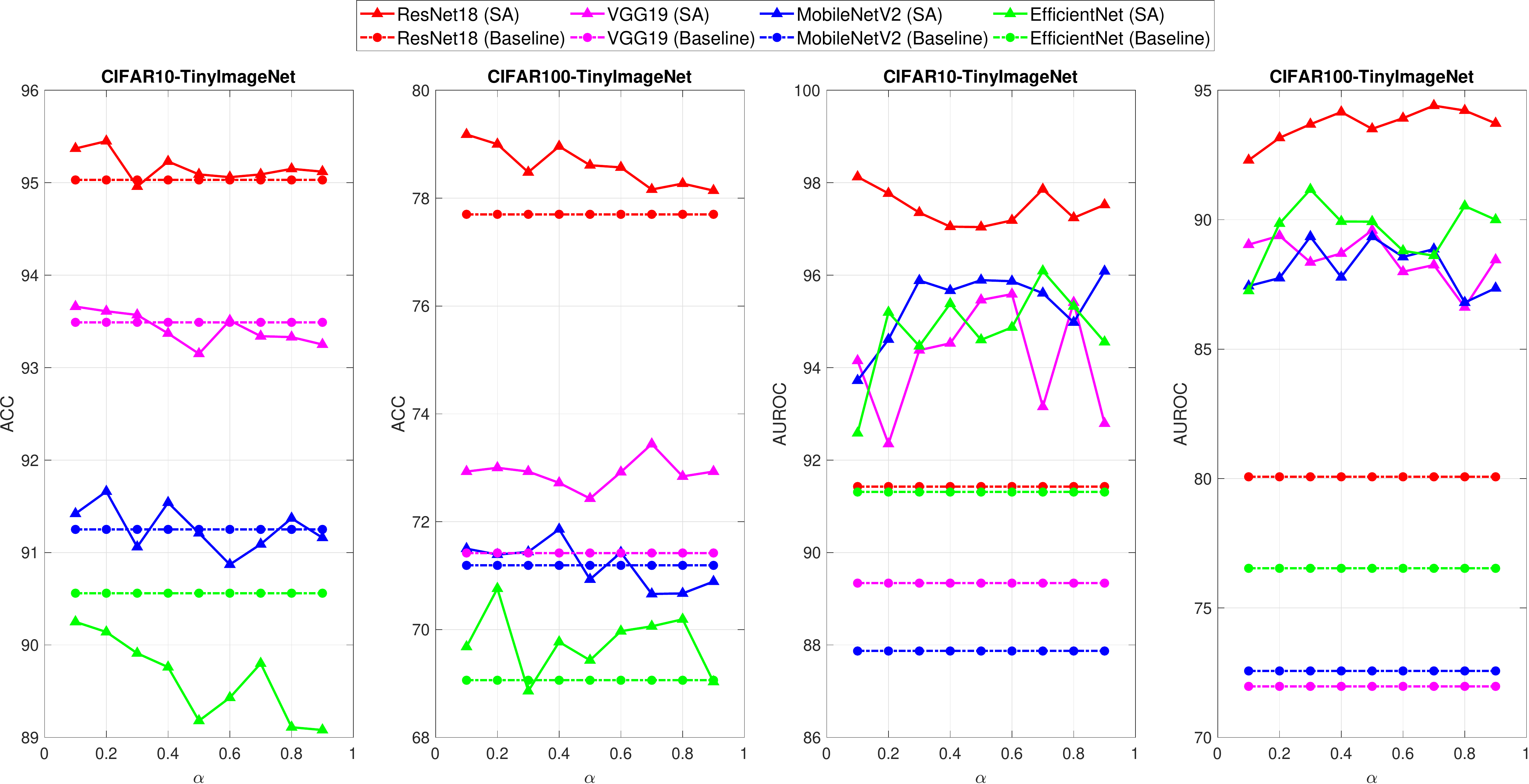}
  \caption{Effect of combination parameter $\alpha$ when the component parameter $\epsilon = 0.05$. Each title contains the information about the training ID dataset and the network architecture. Solid  and dotted lines correspond to the results of SA and the baseline method, respectively. Each point in the first two sub-figures indicates an ACC score on the test dataset corresponding to the training ID dataset, and each point in the last two sub-figures indicates an average AUROC score on the eleven OOD datasets.}
  \label{fig:alpha}
\end{figure*}

\begin{table}[]  %\small
  \renewcommand{\arraystretch}{1.3}
  \renewcommand\tabcolsep{2.0pt}
  \centering
  \caption{Effects of training OOD datasets. T, L and C represent TinyImageNet, LSUN and CelebA, respectively.}
  \label{tb:to}
\begin{tabular}{cccc} \toprule
\multirow{2}{*}{\begin{tabular}[c]{@{}c@{}}In-dist\\ Measure\end{tabular}} & \multirow{2}{*}{Network} & \multirow{2}{*}{Baseline} & SA         \\ \cline{4-4}
                                                                           &                          &                           & T / L / C \\ \midrule\midrule
\multirow{4}{*}{\begin{tabular}[c]{@{}c@{}}CIFAR10 \\ AUROC\end{tabular}}
& ResNet18      & 91.4                      & 98.1 / 98.9 / 96.1 \\
& VGG19          & 89.3                      & 92.4 / 96.4 / 90.7 \\
& MobileNetV2  & 87.9                      & 94.6 / 97.6 / 92.8 \\
& EfficientNet    & 91.3                      & 95.2 / 97.1 / 94.5 \\ \midrule\midrule
\multirow{4}{*}{\begin{tabular}[c]{@{}c@{}}CIFAR10 \\ ACC\end{tabular}}
& ResNet18       & 95.0                      & 95.1 / 95.1 / 95.3           \\
& VGG19           & 93.5                      & 93.6 / 93.7 / 93.5 \\
& MobileNetV2   & 91.3                      & 91.7 / 91.6 / 91.7 \\
& EfficientNet     & 90.6                      & 90.1 / 90.7 / 90.3 \\ \midrule\midrule
\multirow{4}{*}{\begin{tabular}[c]{@{}c@{}}CIFAR100 \\ AUROC\end{tabular}}
& ResNet18        & 80.1                      & 93.0 / 96.4 / 92.1 \\
& VGG19            & 72.0                      & 89.4 / 94.4 / 86.9 \\
& MobileNetV2    & 72.6                      & 87.7 / 92.8 / 83.8 \\
& EfficientNet      & 74.2                      & 89.9 / 94.1 / 86.5 \\ \midrule\midrule
\multirow{4}{*}{\begin{tabular}[c]{@{}c@{}}CIFAR100 \\ ACC\end{tabular}}
& ResNet18          & 77.7                      & 78.6 / 78.6 / 79.0 \\
& VGG19              & 71.4                      & 73.0 / 72.7 / 72.3 \\
& MobileNetV2      & 71.2                      & 71.4 / 72.2 / 71.1 \\
& EfficientNet        & 69.1                      & 70.8 / 69.6 / 70.3 \\ \bottomrule
\end{tabular}
\end{table}

\begin{table}
  \renewcommand{\arraystretch}{1.3}
  \renewcommand\tabcolsep{2.0pt}
  \centering
  \caption{Throughput (image / s) of the baseline and SA methods. Larger values represent more efficient methods.}
  \label{tb:tp}
\begin{tabular}{ccccc}
\toprule
In-dist  & $\#$  ID classes       & Baseline & \begin{tabular}[c]{@{}c@{}}SA \\ ($\epsilon = 0.01$)\end{tabular} & \begin{tabular}[c]{@{}c@{}}SA \\ ($\epsilon = 0.1$)\end{tabular}\\ \midrule
CIFAR10  & 10   & $1.70 \times 10^6$ & $ 0.51 \times 10^6$ & $ 0.51 \times 10^6$\\
CIFAR100 & 100 & $1.69 \times 10^6$ & $ 0.50 \times 10^6$& $ 0.50 \times 10^6$\\ \bottomrule
\end{tabular}
\end{table}

\begin{figure}
  \centering
  \includegraphics[width=0.49\textwidth]{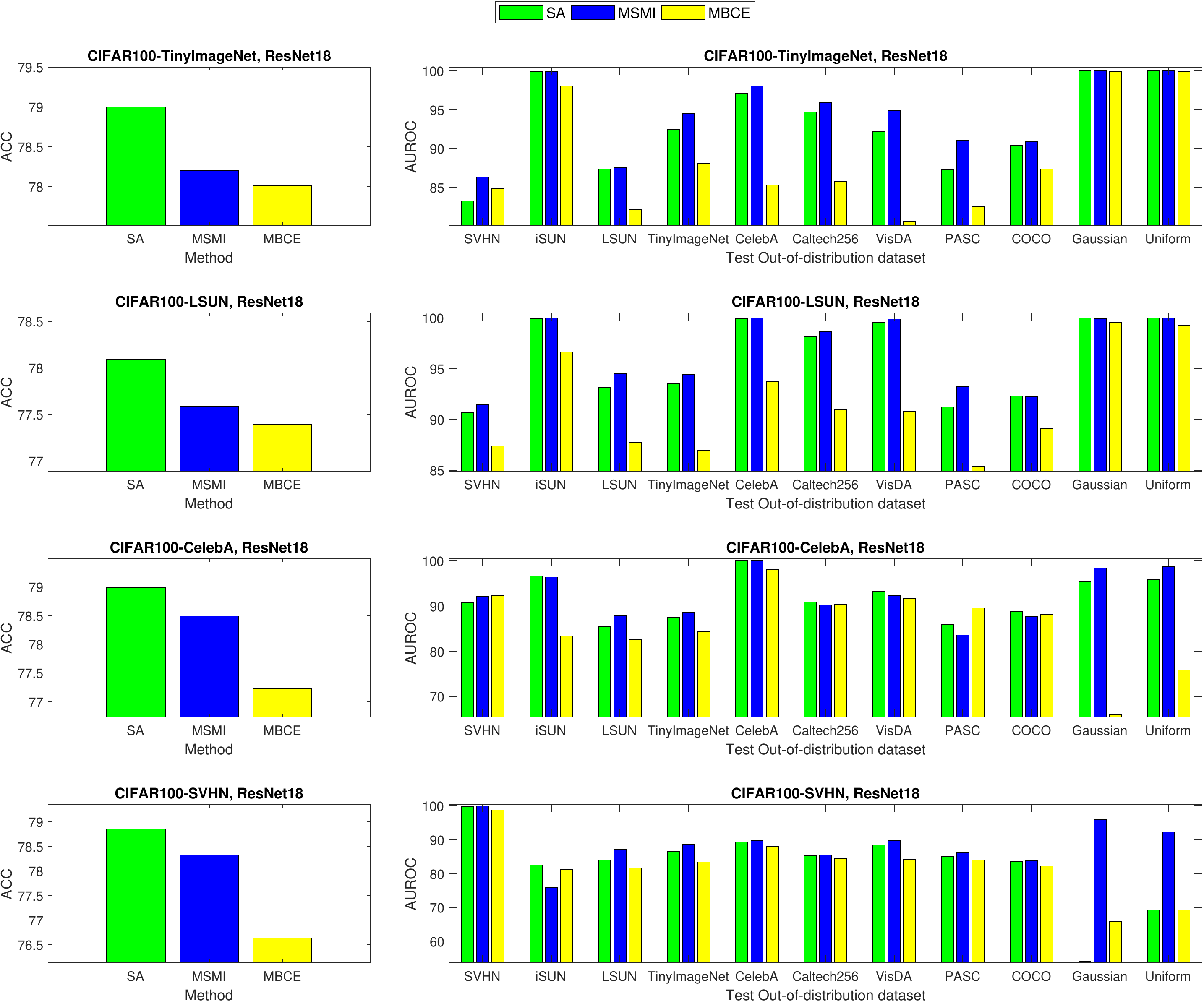}
  \caption{Results of the ablation study on the CIFAR100 dataset. Each title contains the information about the training ID dataset and the network architecture. Each bar in the sub-figures on the left indicates the classification accuracy on the corresponding test ID dataset, and each bar in the sub-figures on the right indicates the detection performance for an OOD dataset. Higher bars are better.}
  \label{fig:abalation}
\end{figure}

\begin{figure*}
  \centering
  \includegraphics[width=1\textwidth]{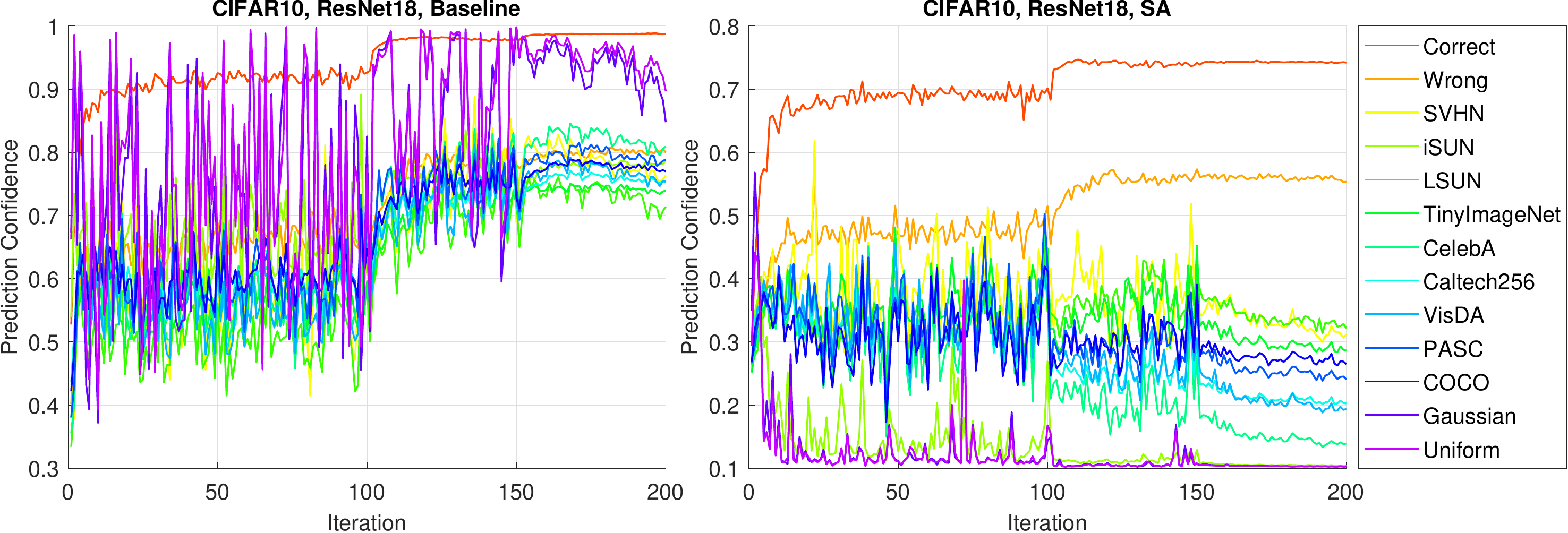}
  \caption{Prediction confidence of ID and OOD test samples from the baseline and SA trained by ResNet18 on CIFAR10. Each solid line represents the change of average prediction confidence of samples from the corresponding test set over time. In the legend, `Correct' and `Wrong' indicate correctly and wrongly classified samples, respectively, and the other names denote different OOD samples.}
  \label{fig:confidence}
\end{figure*}

\begin{figure}
  \centering
\subfigure{
  \includegraphics[width=0.10\textwidth]{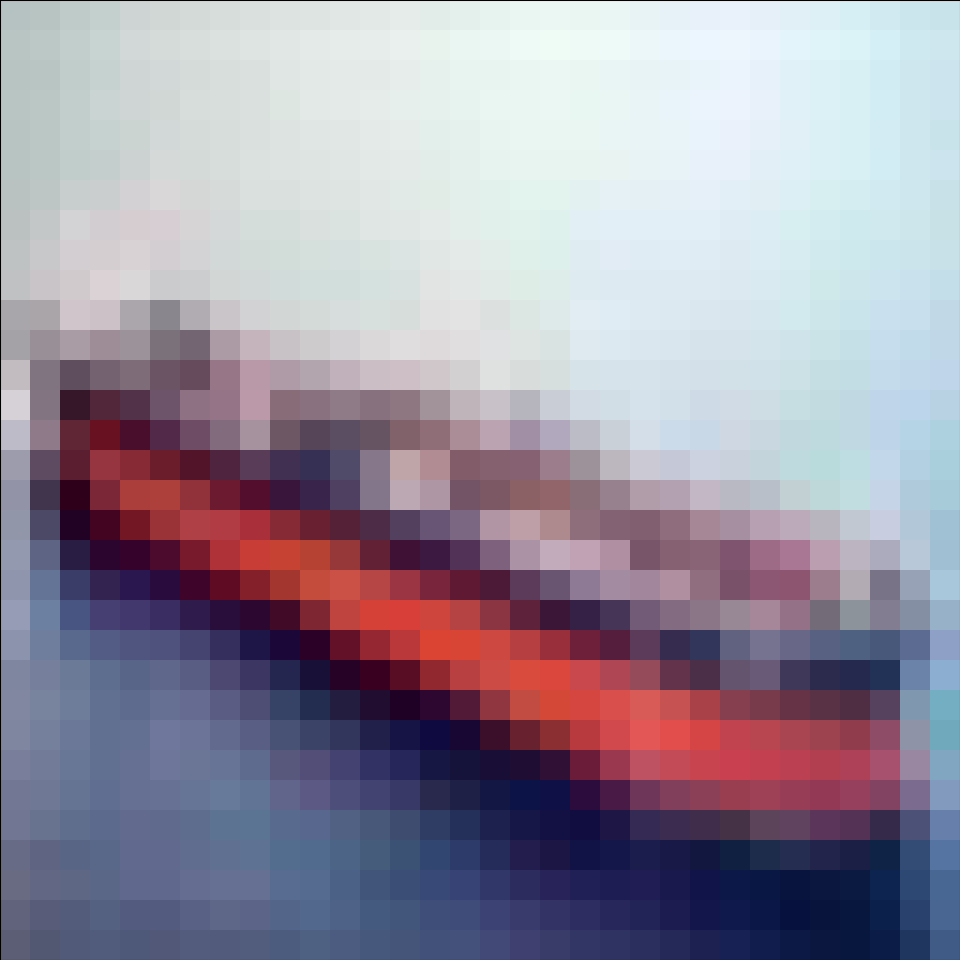}
  \includegraphics[width=0.18\textwidth]{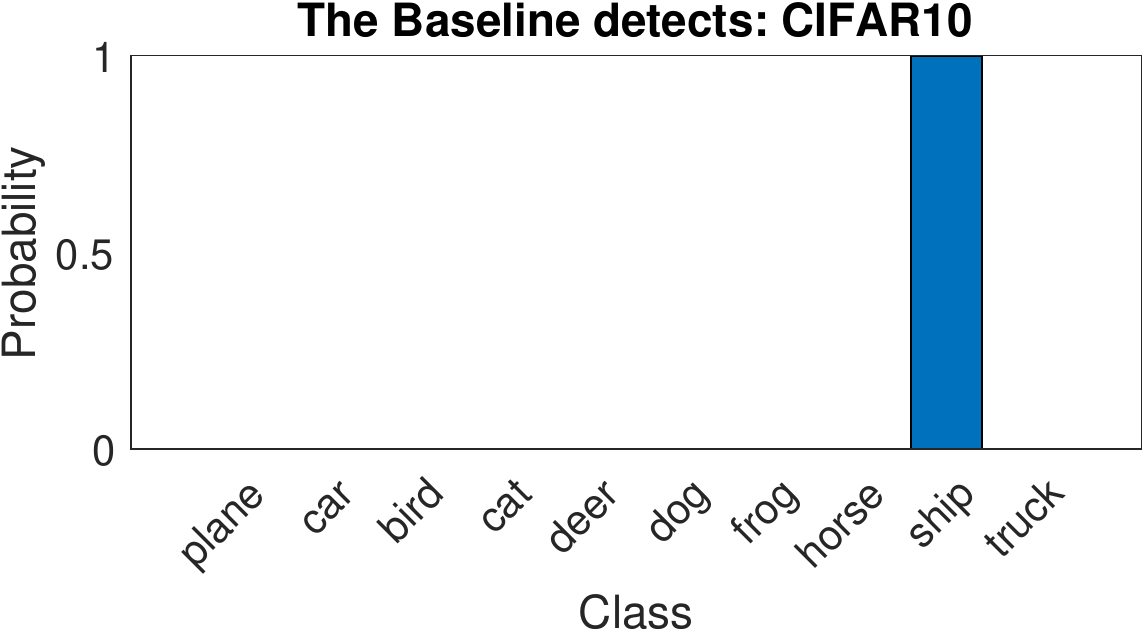}
  \includegraphics[width=0.18\textwidth]{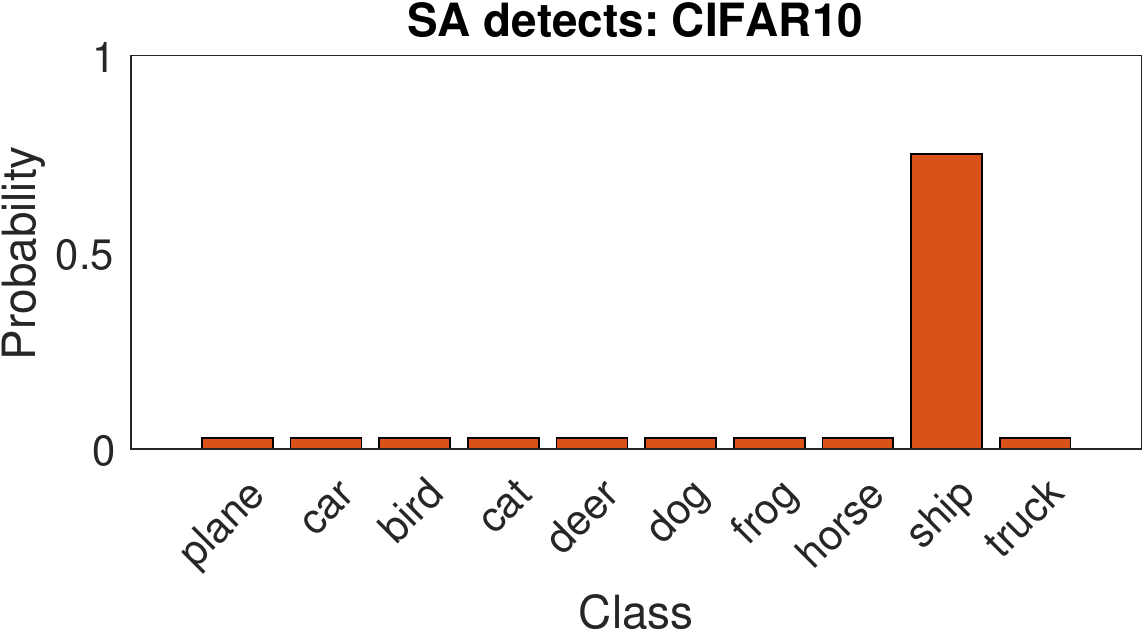}
  }
\subfigure{
  \includegraphics[width=0.10\textwidth]{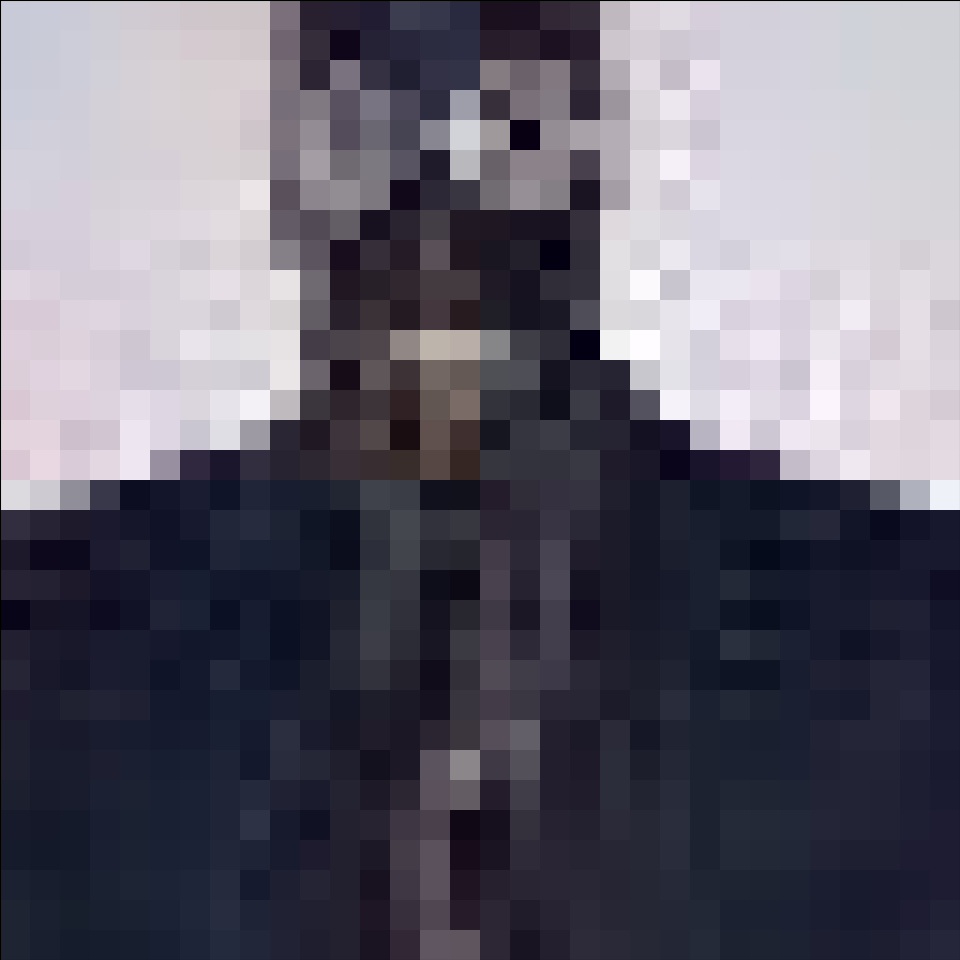}
  \includegraphics[width=0.18\textwidth]{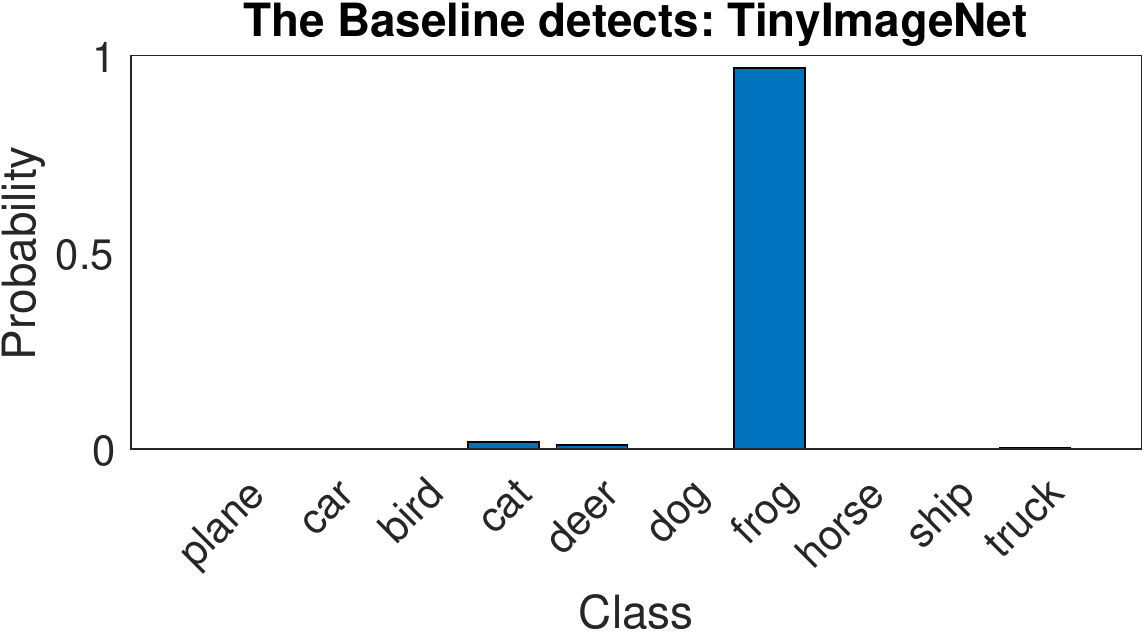}
  \includegraphics[width=0.18\textwidth]{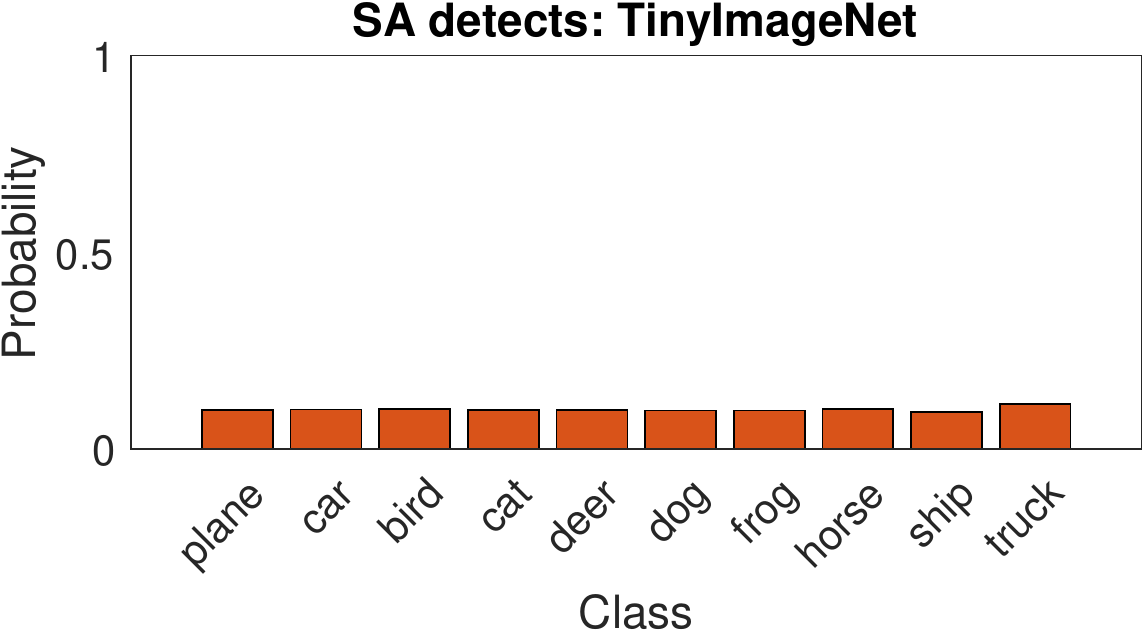}
  }
\subfigure{
  \includegraphics[width=0.10\textwidth]{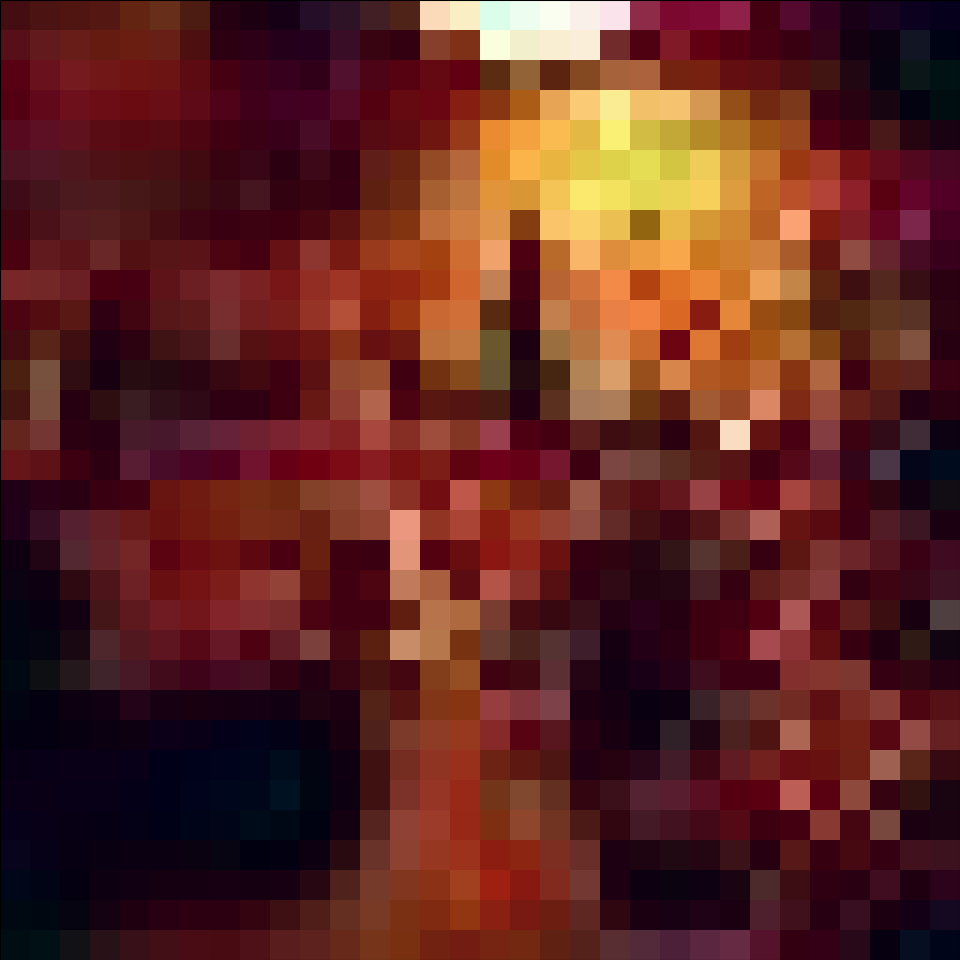}
  \includegraphics[width=0.18\textwidth]{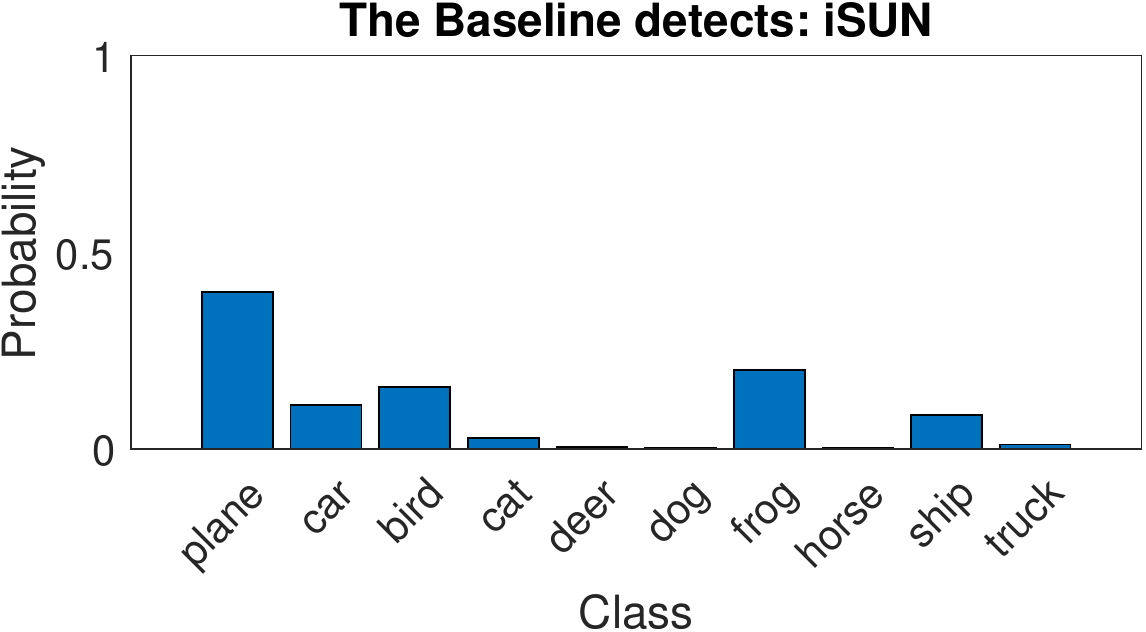}
  \includegraphics[width=0.18\textwidth]{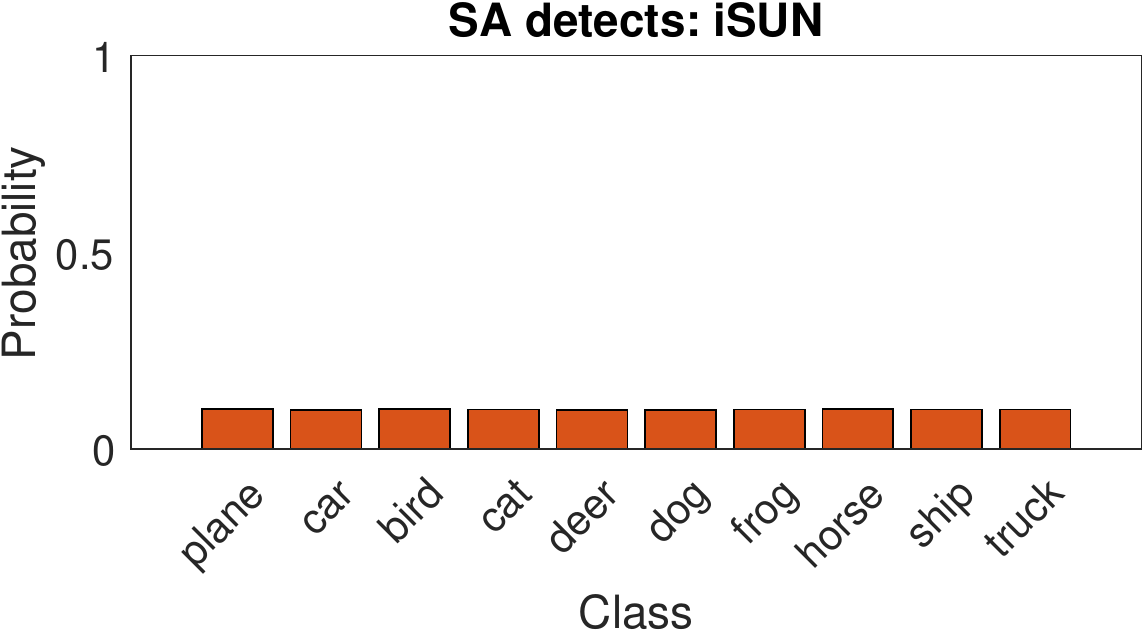}
  }
\subfigure{
  \includegraphics[width=0.10\textwidth]{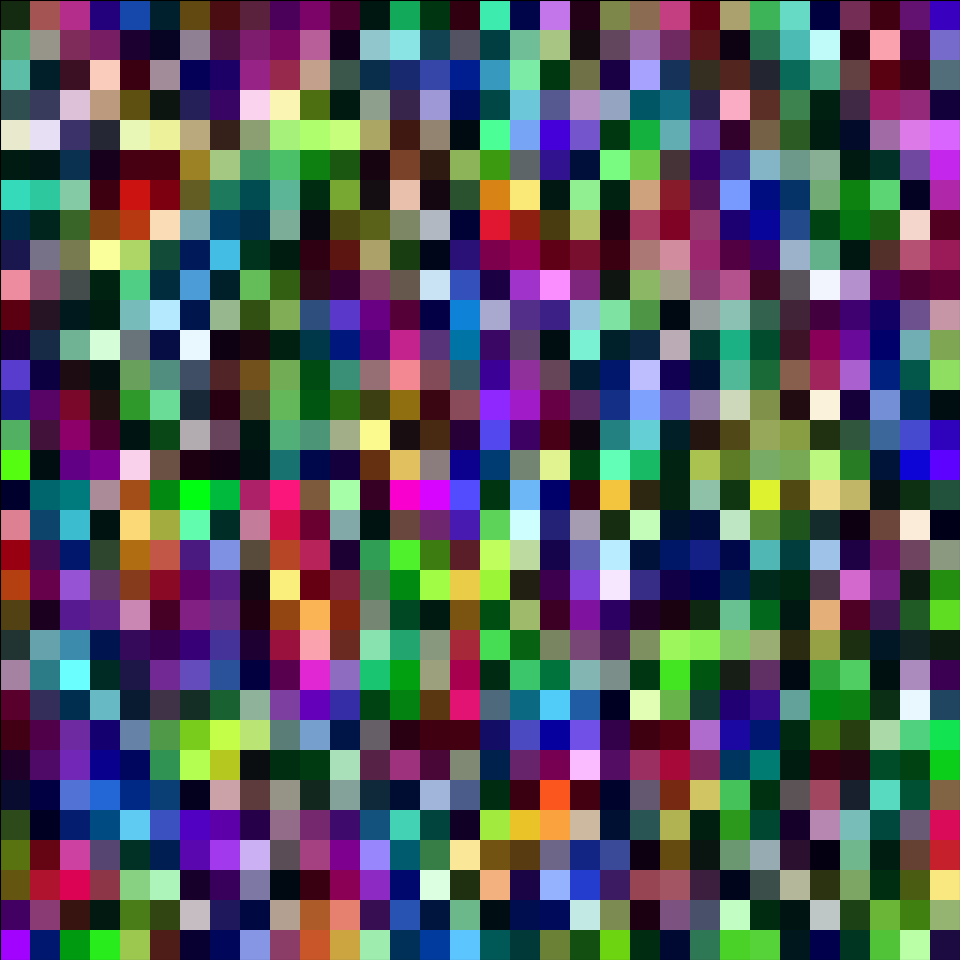}
  \includegraphics[width=0.18\textwidth]{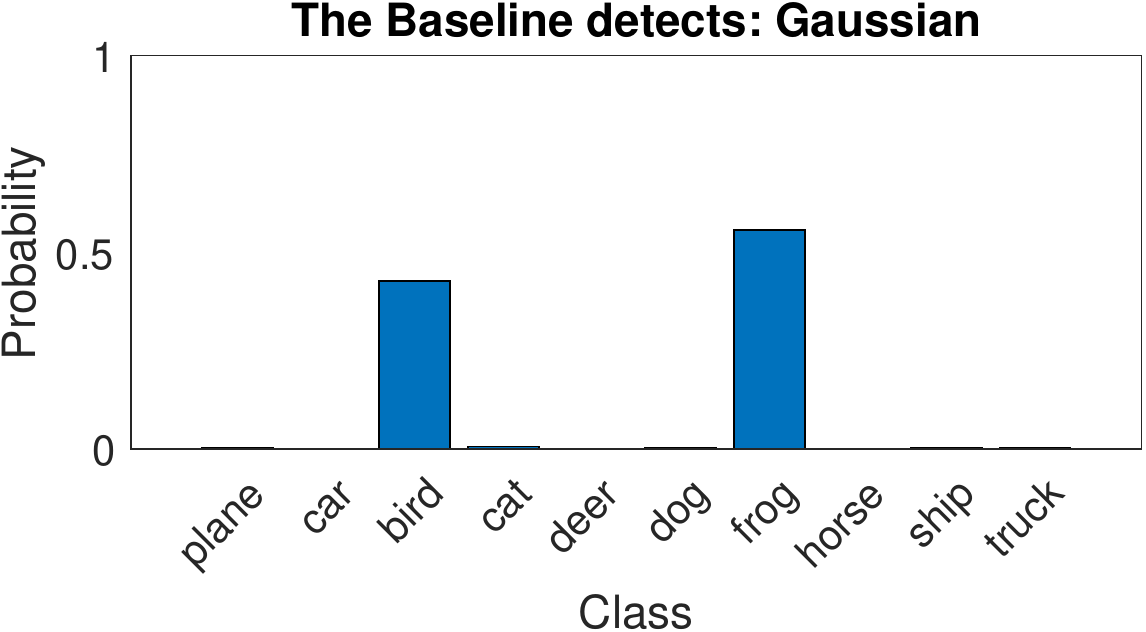}
  \includegraphics[width=0.18\textwidth]{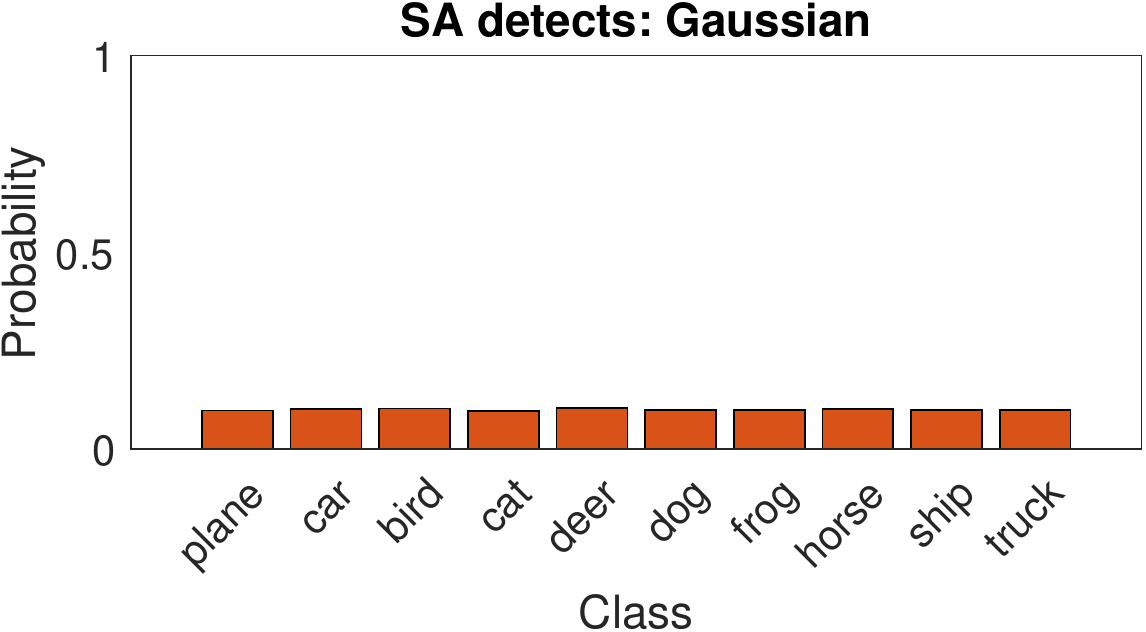}
  }
  \caption{Inputs from different test datasets and their corresponding prediction probabilities of the baseline and SA. The two models are trained by ResNet18 on CIFAR10, and the extra OOD Dataset used in SA is TinyImageNet. The blue and orange bars represent the results of the baseline and SA, respectively.}
  \label{fig:prob}
\end{figure}

\subsubsection{Effect of Combination Parameters}
We assess the impact of the combination parameter $\alpha$ by selecting it from nine evenly-spaced values, ranging from $0.1$ to $0.9$, with a fixed $\epsilon$ of $0.05$. We benchmark SA against the baseline method across four distinct neural network architectures and two ID datasets, considering both ACC and AUROC metrics. These results are compiled in \figurename~\ref{fig:alpha}. Our observations reveal that an $\alpha$ value within the range of $[0.1,0.2]$ generally enhances ACC over the baseline across most ID dataset and network architecture combinations. However, we notice an exception for EfficientNet on the CIFAR10 dataset. This can be attributed to the challenges of boosting the generalization capacity of lightweight, efficiency-optimized models like EfficientNet on relatively simple datasets such as CIFAR10~\cite{EN:19}. Regarding AUROC, SA consistently outstrips the baseline across all choices of $\alpha$, substantiating the validity and effectiveness of leveraging OOD samples to heighten network sensitivity towards such samples. Although SA reaches the peak OOD detection performance at around $\alpha = 0.5$, a smaller $\alpha$ emerges as a more balanced choice for simultaneously maintaining high OOD detection capabilities while also improving classification accuracy relative to the baseline.

\subsection{Effect of Training OOD Datasets}~\label{sec:eop}
To examine the impact of different training OOD datasets, we utilize three distinct OOD datasets for training on the CIFAR10 ID dataset: TinyImageNet (T), LSUN (L), and CelebA (C). We run experiments using four neural network architectures and report the ACC and average AUROC results across eleven OOD datasets in \tablename~\ref{tb:to}. We find that the SA method, when trained on different OOD datasets, consistently outperforms the baseline in terms of both ACC and AUROC. Moreover, different training OOD datasets yield similar ID classification performance but only slightly different OOD detection results. This implies that SA effectively minimizes the impact of the incorporated training OOD samples on ID classification by leveraging adaptive supervision information. The minor variations in OOD detection can be attributed to the natural distribution discrepancies between the training and test OOD samples.

\begin{figure*}
\centering
    \includegraphics[width=1\textwidth]{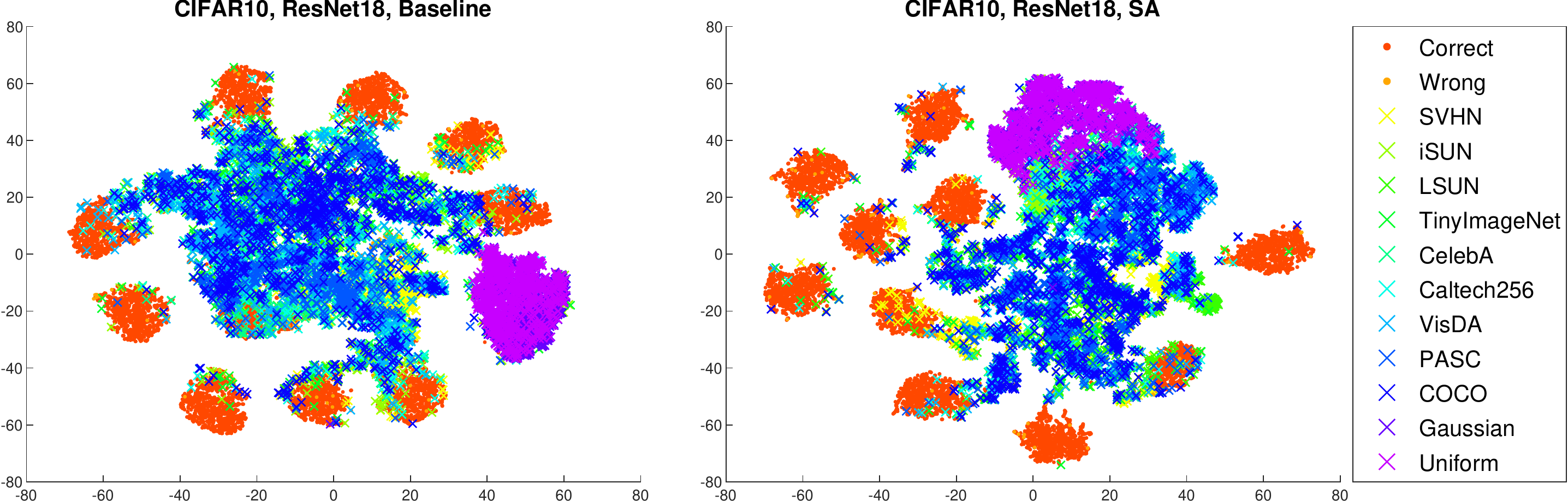}
  \caption{t-SNE visualization of ResNet18 features from the baseline and SA trained on CIFAR10. The color of points indicates the datasets of the corresponding samples. `Correct' and `Wrong' indicate correctly and wrongly classified samples, respectively. The other names denote different OOD samples.}
  \label{fig:embedding}
\end{figure*}

\subsection{Ablation Study}
SA comprises two main components: MSMI and MBCE. MSMI outlines the type of supervision information needed for OOD samples, while MBCE calculates this adaptive supervision information for training OOD-sensitive networks. To validate the necessity and complementarity of both MSMI and MBCE, we conduct a series of ablation experiments using ResNet18 on the CIFAR100 dataset. For training, the OOD datasets included are TinyImageNet, LSUN, CelebA, and SVHN. In these experiments, MSMI aims to optimize the objective function denoted by Eq.~(\ref{eq:msmi}). Here, the adaptive supervision information $P(y|x)$ is directly approximated by a parametric discriminator $Q_{\theta}(y|\mathbf{x})$, which employs the standard softmax function. On the other hand, MBCE works to optimize another objective function represented by Eq.~(\ref{eq:mbce}). It learns $\mathcal{D}(\mathbf{x},y)$ and subsequently infers the discriminator $Q_{\theta}(y|\mathbf{x})$ as defined by Eq.~(\ref{eq:Q3}). For a fair comparison, and based on the recommendations outlined in Section~\ref{sec:eop}, we uniformly set the component parameter $\epsilon = 0.05$ for all tested methods, and for SA, the combination parameter $\alpha = 0.2$ is applied.

\figurename~\ref{fig:abalation} presents the ACC and AUROC outcomes for the methods SA, MSMI, and MBCE, leading to several notable observations. While the networks trained solely using MBCE demonstrate limited classification accuracy, incorporating MSMI into MBCE to create the SA method results in significant improvements. This underscores the importance of MBCE in capturing the data relationships between ID and OOD samples, thereby enhancing the quality of the supervision information estimated by MSMI. When it comes to the OOD detection performance, MSMI emerges as the most effective, highlighting the utility of adaptive supervision information in identifying OOD samples. Conversely, MBCE exhibits a diminished OOD detection capability, which suggests that merely segregating ID and OOD samples, without taking into account the relationships between samples and their corresponding labels, is inadequate for training OOD-sensitive networks. By integrating the strengths of both MSMI and MBCE, SA retains a similar level of OOD detection performance as MSMI, while also boosting the classification accuracy through MBCE, albeit with a negligible loss in OOD detection performance. Therefore, we conclude that MSMI and MBCE are complementary components, collectively enabling the balanced performance of both ID generalization and OOD detection.

\subsection{Efficiency Analysis}
The throughput of both the baseline and SA methods is presented in \tablename~\ref{tb:tp} to assess their efficiency. We find that the efficiency of SA is only loosely correlated with the number of ID classes and the composition of OOD samples. Moreover, SA operates at approximately one-third the efficiency of the baseline method. This drop in efficiency is largely attributable to the second term in the objective function Eq.~(\ref{eq:SA}), which aims to learn adaptive supervision information from the OOD samples. In contrast, optimizing the efficiency of the first term is essentially the same as optimizing the cross-entropy loss when training the pre-trained network in the baseline method. Therefore, while SA does incur an efficiency cost, operating at around two-thirds less efficiency compared to the baseline, it compensates for this by delivering state-of-the-art OOD detection performance along with competitive classification accuracy. This presents a calculated trade-off between learning effectiveness and computational efficiency.

\subsection{Visualization of ID and OOD Separability}
To offer an intuitive sense of how effectively the SA method differentiates between ID and OOD samples, we present qualitative analyses focusing on the prediction confidence trends, the ultimate prediction probabilities, and the feature distributions. We utilize the CIFAR10 dataset as the ID dataset and run both the baseline and SA methods on the ResNet18 architecture. For the SA method, TinyImageNet is used as the additional OOD dataset, with experimental setups aligned with those described in Section~\ref{sec:comRes}.

\subsubsection{Prediction Confidence Trend}
The trends in prediction confidence for various test datasets are illustrated in \figurename~\ref{fig:confidence}. For both the baseline and SA methods, we compute the average prediction confidence at each iteration for each test dataset, taking into account both correctly and incorrectly classified samples in the test ID dataset. As the number of iterations increases, the average prediction confidence for both correctly and incorrectly classified samples in SA gradually rises, whereas it exhibits a declining trend for all the test OOD datasets. In contrast, the baseline method shows an increase in average prediction confidence for test OOD datasets, similar to what is observed for the test ID dataset. Consequently, the widening gap in average prediction confidence between ID and OOD samples renders SA more adept at distinguishing between these two types of samples.

\subsubsection{Prediction Probability}
The final prediction probabilities for individual samples are presented in \figurename~\ref{fig:prob}. Here, we showcase four test samples originating from different datasets, along with their corresponding prediction probabilities as generated by both the baseline and SA methods. Among these samples, those from the CIFAR10 test set are considered ID, while the others are categorized as OOD. It is evident that both the baseline and SA exhibit high confidence for the correct class when dealing with an ID sample. However, SA demonstrates a greater sensitivity to OOD samples, as it significantly suppresses the confidence levels across all classes for the three OOD samples, in contrast to the higher confidences exhibited by the baseline.

\subsubsection{Feature Distribution}
The feature distributions for both the baseline and SA are visualized using t-SNE~\cite{SNE:08} embeddings in \figurename~\ref{fig:embedding}. It is important to note that the CIFAR10 dataset consists of 10 classes. In the case of the baseline method, only $9$ classes can be clearly delineated, with the remaining class region exhibiting significant overlap with OOD samples. Conversely, for SA, all classes are clearly discernible, as they are well separated and distinct from the OOD samples. This underscores the enhanced ability of SA to effectively differentiate between ID and OOD samples.

\section{Conclusions and Future Work}\label{sec:conclusion}
Addressing the non-IID complexity and distributional vulnerability inherent in deep neural networks presents a significant challenge, especially when test samples are OOD differing from ID training samples. This challenge demands a balanced dual ability of maintaining ID classification accuracy while enhancing OOD detection capabilities. To address this challenge, we introduce Supervision Adaptation (SA), a comprehensive information-theoretic framework aimed at fortifying OOD detection without compromising ID classification performance. SA tackles the issue of crafting appropriate supervision information for OOD samples in a way that adapts them to ID samples. It initially uncovers the nature of this supervision information by leveraging the mutual information between ID samples and their labels, within a data space that includes both ID and OOD samples. Subsequently, SA approximates this supervision information by solving a series of binary classification problems that scrutinize the correlations between ID and OOD samples. Through empirical evaluations across diverse neural architectures and datasets, SA demonstrates superiority over state-of-the-art methods in various domains, including OOD detection, OOD generalization, domain adaptation, and generalization improvement. Moreover, SA delivers competitive ID classification performance, consistently outpacing the baseline method. Hence, SA achieves the desired equilibrium between OOD sensitivity and ID classification efficiency.

The SA framework offers several avenues for future expansion, including but not limited to: accommodating the open set assumption, integrating non-stationary ID or OOD samples, and generating specific OOD samples for a pretrained network when training OOD samples are not readily available. Thus, SA sets a new benchmark for addressing the intricate balance between ID classification and OOD detection in deep neural networks.

% or


% \clearpage

\ifCLASSOPTIONcompsoc
  \section*{Acknowledgments}
\else
  \section*{Acknowledgment}
\fi

This work was supported in part by the Australian Research Council Discovery under Grant DP190101079 and in part by the Future Fellowship under Grant FT190100734.

\ifCLASSOPTIONcaptionsoff
  \newpage
\fi



%\bibliographystyle{IEEEtran}

\bibliographystyle{IEEEtran}
\bibliography{ref}

\begin{thebibliography}{10}
\providecommand{\url}[1]{#1}
\csname url@samestyle\endcsname
\providecommand{\newblock}{\relax}
\providecommand{\bibinfo}[2]{#2}
\providecommand{\BIBentrySTDinterwordspacing}{\spaceskip=0pt\relax}
\providecommand{\BIBentryALTinterwordstretchfactor}{4}
\providecommand{\BIBentryALTinterwordspacing}{\spaceskip=\fontdimen2\font plus
\BIBentryALTinterwordstretchfactor\fontdimen3\font minus
  \fontdimen4\font\relax}
\providecommand{\BIBforeignlanguage}[2]{{%
\expandafter\ifx\csname l@#1\endcsname\relax
\typeout{** WARNING: IEEEtran.bst: No hyphenation pattern has been}%
\typeout{** loaded for the language `#1'. Using the pattern for}%
\typeout{** the default language instead.}%
\else
\language=\csname l@#1\endcsname
\fi
#2}}
\providecommand{\BIBdecl}{\relax}
\BIBdecl

\bibitem{GE:17}
C.~Zhang, S.~Bengio, M.~Hardt, B.~Recht, and O.~Vinyals, ``Understanding deep
  learning requires rethinking generalization,'' in \emph{5th International
  Conference on Learning Representations}, 2017, pp. 1--15.

\bibitem{CAL:17}
C.~Guo, G.~Pleiss, Y.~Sun, and K.~Q. Weinberger, ``On calibration of modern
  neural networks,'' in \emph{34th International Conference on Machine
  Learning}, 2017, pp. 1321--1330.

\bibitem{C22beyiid}
L.~Cao, ``Beyond i.i.d.: Non-iid thinking, informatics, and learning,''
  \emph{{IEEE} Intell. Syst.}, vol.~37, no.~4, pp. 5--17, 2022.

\bibitem{CaoY022}
L.~Cao, P.~S. Yu, and Z.~Zhao, ``Shallow and deep non-iid learning on complex
  data,'' in \emph{The 28th {ACM} {SIGKDD} Conference on Knowledge Discovery
  and Data Mining}, 2022, pp. 4774--4775.

\bibitem{Z22ood}
Z.~Zhao, L.~Cao, and K.-Y. Lin, ``Revealing the distributional vulnerability of
  discriminators by implicit generators,'' \emph{{IEEE} Trans. Pattern Anal.
  Mach. Intell.}, vol.~45, no.~7, pp. 8888--8901, 2023.

\bibitem{UN:17}
A.~Kendall and Y.~Gal, ``What uncertainties do we need in bayesian deep
  learning for computer vision?'' in \emph{Advances in Neural Information
  Processing Systems 30}, 2017, pp. 5574--5584.

\bibitem{AD:15}
I.~J. Goodfellow, J.~Shlens, and C.~Szegedy, ``Explaining and harnessing
  adversarial examples,'' in \emph{3rd International Conference on Learning
  Representations}, 2015, pp. 1--11.

\bibitem{DP:16}
Y.~Gal and Z.~Ghahramani, ``Dropout as a bayesian approximation: Representing
  model uncertainty in deep learning,'' in \emph{33nd International Conference
  on Machine Learning}, 2016, pp. 1050--1059.

\bibitem{OS:16}
A.~Bendale and T.~E. Boult, ``Towards open set deep networks,'' in
  \emph{Proceedings of the {IEEE} Conference on Computer Vision and Pattern
  Recognition}, 2016, pp. 1563--1572.

\bibitem{PN:18}
A.~Malinin and M.~J.~F. Gales, ``Predictive uncertainty estimation via prior
  networks,'' in \emph{Advances in Neural Information Processing Systems 31},
  2018, pp. 7047--7058.

\bibitem{GE:03}
K.~Vladimir, D.~Panchenko, and F.~Lozano, ``Bounding the generalization error
  of convex combinations of classifiers: balancing the dimensionality and the
  margins,'' \emph{The Annals of Applied Probability}, vol.~13, no.~1, pp.
  213--252, 2003.

\bibitem{VB:19}
B.~Poole, S.~Ozair, A.~van~den Oord, A.~Alemi, and G.~Tucker, ``On variational
  bounds of mutual information,'' in \emph{36th International Conference on
  Machine Learning}, 2019, pp. 5171--5180.

\bibitem{GOOD:21}
J.~Yang, K.~Zhou, Y.~Li, and Z.~Liu, ``Generalized out-of-distribution
  detection: {A} survey,'' \emph{CoRR}, vol. abs/2110.11334, 2021.

\bibitem{OOD:21}
M.~Salehi, H.~Mirzaei, D.~Hendrycks, Y.~Li, M.~H. Rohban, and M.~Sabokrou, ``A
  unified survey on anomaly, novelty, open-set, and out of-distribution
  detection: Solutions and future challenges,'' \emph{Trans. Mach. Learn.
  Res.}, vol. 2022, 2022.

\bibitem{GUD:21}
J.~Wang, C.~Lan, C.~Liu, Y.~Ouyang, and T.~Qin, ``Generalizing to unseen
  domains: {A} survey on domain generalization,'' \emph{Proceedings of the
  Thirtieth International Joint Conference on Artificial Intelligence}, pp.
  4627--4635, 2021.

\bibitem{DG:23}
K.~Zhou, Z.~Liu, Y.~Qiao, T.~Xiang, and C.~C. Loy, ``Domain generalization: {A}
  survey,'' \emph{{IEEE} Trans. Pattern Anal. Mach. Intell.}, vol.~45, no.~4,
  pp. 4396--4415, 2023.

\bibitem{ML:14}
S.~Shalev-Shwartz and S.~Ben-David, \emph{Understanding Machine Learning From
  Theory to Algorithms}.\hskip 1em plus 0.5em minus 0.4em\relax Cambridge
  University Press, 2014.

\bibitem{BL:17}
D.~Hendrycks and K.~Gimpel, ``A baseline for detecting misclassified and
  out-of-distribution examples in neural networks,'' in \emph{5th International
  Conference on Learning Representations}, 2017, pp. 1--12.

\bibitem{DRF:20}
E.~Zisselman and A.~Tamar, ``Deep residual flow for out of distribution
  detection,'' in \emph{Proceedings of the {IEEE} Conference on Computer Vision
  and Pattern Recognition}, 2020, pp. 13\,991--14\,000.

\bibitem{EB:20}
W.~Liu, X.~Wang, J.~D. Owens, and Y.~Li, ``Energy-based out-of-distribution
  detection,'' in \emph{Advances in Neural Information Processing Systems 33},
  no. 1--12, 2020.

\bibitem{ODIN:18}
S.~Liang, Y.~Li, and R.~Srikant, ``Enhancing the reliability of
  out-of-distribution image detection in neural networks,'' in \emph{6th
  International Conference on Learning Representations}, 2018, pp. 1--27.

\bibitem{GO:18}
K.~Lee, H.~Lee, K.~Lee, and J.~Shin, ``Training confidence-calibrated
  classifiers for detecting out-of-distribution samples,'' in \emph{6th
  International Conference on Learning Representations}, 2018, pp. 1--16.

\bibitem{LDA:03}
D.~M. Blei, A.~Y. Ng, and M.~I. Jordan, ``Latent dirichlet allocation,''
  \emph{J. Mach. Learn. Res.}, vol.~3, pp. 993--1022, 2003.

\bibitem{OE:19}
D.~Hendrycks, M.~Mazeika, and T.~G. Dietterich, ``Deep anomaly detection with
  outlier exposure,'' in \emph{7th International Conference on Learning
  Representations}, 2019, pp. 1--18.

\bibitem{ADB:19}
M.~Hein, M.~Andriushchenko, and J.~Bitterwolf, ``Why relu networks yield
  high-confidence predictions far away from the training data and how to
  mitigate the problem,'' in \emph{Proceedings of the {IEEE} Conference on
  Computer Vision and Pattern Recognition}, 2019, pp. 41--50.

\bibitem{ESS:18}
A.~Vyas, N.~Jammalamadaka, X.~Zhu, D.~Das, B.~Kaul, and T.~L. Willke,
  ``Out-of-distribution detection using an ensemble of self supervised
  leave-out classifiers,'' in \emph{Proceedings of the European Conference on
  Computer Vision}, 2018, pp. 560--574.

\bibitem{DCC:20}
Y.~Hsu, Y.~Shen, H.~Jin, and Z.~Kira, ``Generalized {ODIN} detecting
  out-of-distribution image without learning from out-of-distribution data,''
  in \emph{Proceedings of the {IEEE} Conference on Computer Vision and Pattern
  Recognition}, 2020, pp. 10\,948--10\,957.

\bibitem{MOS:21}
R.~Huang and Y.~Li, ``{MOS:} towards scaling out-of-distribution detection for
  large semantic space,'' in \emph{Proceedings of the {IEEE} Conference on
  Computer Vision and Pattern Recognition}, 2021, pp. 8710--8719.

\bibitem{SSD:21}
V.~Sehwag, M.~Chiang, and P.~Mittal, ``{SSD:} {A} unified framework for
  self-supervised outlier detection,'' in \emph{9th International Conference on
  Learning Representations}, 2021, pp. 1--17.

\bibitem{RSC:20}
Z.~Huang, H.~Wang, E.~P. Xing, and D.~Huang, ``Self-challenging improves
  cross-domain generalization,'' in \emph{Proceedings of the European
  Conference on Computer Vision}, 2020, pp. 124--140.

\bibitem{IRM:19}
M.~Arjovsky, L.~Bottou, I.~Gulrajani, and D.~Lopez{-}Paz, ``Invariant risk
  minimization,'' \emph{CoRR}, vol. abs/1907.02893, 2019.

\bibitem{AND:21}
G.~Parascandolo, A.~Neitz, A.~Orvieto, L.~Gresele, and B.~Sch{\"{o}}lkopf,
  ``Learning explanations that are hard to vary,'' in \emph{9th International
  Conference on Learning Representations}, 2021, pp. 1--24.

\bibitem{DRO:20}
S.~Sagawa, P.~W. Koh, T.~B. Hashimoto, and P.~Liang, ``Distributionally robust
  neural networks,'' in \emph{8th International Conference on Learning
  Representations}, 2020, pp. 1--19.

\bibitem{KHRM:21}
J.~Liu, Z.~Hu, P.~Cui, B.~Li, and Z.~Shen, ``Kernelized heterogeneous risk
  minimization,'' in \emph{Advances in Neural Information Processing Systems},
  2021, pp. 21\,720--21\,731.

\bibitem{SAL:21}
J.~Liu, Z.~Shen, P.~Cui, L.~Zhou, K.~Kuang, B.~Li, and Y.~Lin, ``Stable
  adversarial learning under distributional shifts,'' in \emph{The Thirty-Fifth
  {AAAI} Conference on Artificial Intelligence}, 2021, pp. 8662--8670.

\bibitem{DIL:22}
J.~Liu, J.~Wu, J.~Peng, Z.~Shen, B.~Li, and P.~Cui, ``Distributionally
  invariant learning: Rationalization and practical algorithms,'' \emph{CoRR},
  vol. abs/2206.02990, 2022.

\bibitem{GNA:23}
X.~Zhang, R.~Xu, H.~Yu, H.~Zou, and P.~Cui, ``Gradient norm aware minimization
  seeks first-order flatness and improves generalization,'' in
  \emph{Proceedings of the {IEEE} Conference on Computer Vision and Pattern
  Recognition}, 2023, pp. 20\,247--20\,257.

\bibitem{CI:22}
K.~Lin, J.~Zhou, Y.~Qiu, and W.~Zheng, ``Adversarial partial domain adaptation
  by cycle inconsistency,'' in \emph{Proceedings of the European Conference on
  Computer Vision}, 2022, pp. 530--548.

\bibitem{KLIEP:07}
M.~Sugiyama, M.~Krauledat, and K.~M{\"{u}}ller, ``Covariate shift adaptation by
  importance weighted cross validation,'' \emph{J. Mach. Learn. Res.}, vol.~8,
  pp. 985--1005, 2007.

\bibitem{DASVM:10}
L.~Bruzzone and M.~Marconcini, ``Domain adaptation problems: {A} {DASVM}
  classification technique and a circular validation strategy,'' \emph{{IEEE}
  Trans. Pattern Anal. Mach. Intell.}, vol.~32, no.~5, pp. 770--787, 2010.

\bibitem{DANN:16}
Y.~Ganin, E.~Ustinova, H.~Ajakan, P.~Germain, H.~Larochelle, F.~Laviolette,
  M.~Marchand, and V.~S. Lempitsky, ``Domain-adversarial training of neural
  networks,'' \emph{J. Mach. Learn. Res.}, vol.~17, pp. 1--59, 2016.

\bibitem{CORAL:16}
B.~Sun and K.~Saenko, ``Deep {CORAL:} correlation alignment for deep domain
  adaptation,'' in \emph{Computer Vision - {ECCV} 2016 Workshops}, 2016, pp.
  443--450.

\bibitem{DA:12}
P.~Y. Simard, Y.~LeCun, J.~S. Denker, and B.~Victorri, ``Transformation
  invariance in pattern recognition - tangent distance and tangent
  propagation,'' \emph{Neural Networks: Tricks of the Trade}, pp. 235--269,
  2012.

\bibitem{FL:15}
K.~Simonyan and A.~Zisserman, ``Very deep convolutional networks for
  large-scale image recognition,'' in \emph{3rd International Conference on
  Learning Representations}, 2015, pp. 1--14.

\bibitem{RE:20}
Z.~Zhong, L.~Zheng, G.~Kang, S.~Li, and Y.~Yang, ``Random erasing data
  augmentation,'' in \emph{The Thirty-Fourth {AAAI} Conference on Artificial
  Intelligence}, 2020, pp. 13\,001--13\,008.

\bibitem{GN:17}
H.~Noh, T.~You, J.~Mun, and B.~Han, ``Regularizing deep neural networks by
  noise: Its interpretation and optimization,'' \emph{Advances in Neural
  Information Processing Systems 30}, pp. 5109--5118, 2017.

\bibitem{MIXUP:18}
H.~Zhang, M.~Ciss{\'{e}}, Y.~N. Dauphin, and D.~Lopez{-}Paz, ``mixup: Beyond
  empirical risk minimization,'' in \emph{6th International Conference on
  Learning Representations}, 2018, pp. 1--13.

\bibitem{CP:17}
G.~Pereyra, G.~Tucker, J.~Chorowski, L.~Kaiser, and G.~E. Hinton,
  ``Regularizing neural networks by penalizing confident output
  distributions,'' in \emph{5th International Conference on Learning
  Representations}, 2017, pp. 1--11.

\bibitem{KD:20}
L.~Yuan, F.~E.~H. Tay, G.~Li, T.~Wang, and J.~Feng, ``Revisiting knowledge
  distillation via label smoothing regularization,'' in \emph{Proceedings of
  the {IEEE} Conference on Computer Vision and Pattern Recognition}, 2020, pp.
  3902--3910.

\bibitem{MINE:18}
M.~I. Belghazi, A.~Baratin, S.~Rajeswar, S.~Ozair, Y.~Bengio, R.~D. Hjelm, and
  A.~C. Courville, ``Mutual information neural estimation,'' in \emph{35th
  International Conference on Machine Learning}.

\bibitem{LB:03}
D.~Barber and F.~V. Agakov, ``The {IM} algorithm: {A} variational approach to
  information maximization,'' in \emph{Advances in Neural Information
  Processing Systems 16}, 2003, pp. 201--208.

\bibitem{RC:02}
P.~L. Bartlett and S.~Mendelson, ``Rademacher and gaussian complexities: Risk
  bounds and structural results,'' \emph{J. Mach. Learn. Res.}, vol.~3, pp.
  463--482, 2002.

\bibitem{DE:12}
M.~Sugiyama, T.~Suzuki, and T.~Kanamori, ``Density-ratio matching under the
  bregman divergence: a unified framework of density-ratio estimation,''
  \emph{Annals of the Institute of Statistical Mathematics}, vol.~64, no.~5,
  pp. 1009--1044, 2012.

\bibitem{EIM:20}
P.~Becker, O.~Arenz, and G.~Neumann, ``Expected information maximization: Using
  the i-projection for mixture density estimation,'' in \emph{8th International
  Conference on Learning Representations}, 2020, pp. 1--16.

\bibitem{PRML:06}
C.~M. Bishop, \emph{Pattern Recognition and Machine Learning}.\hskip 1em plus
  0.5em minus 0.4em\relax springer, 2006.

\bibitem{RES:16}
K.~He, X.~Zhang, S.~Ren, and J.~Sun, ``Deep residual learning for image
  recognition,'' in \emph{Proceedings of the {IEEE} Conference on Computer
  Vision and Pattern Recognition}, 2016, pp. 770--778.

\bibitem{VGG:15}
C.~Szegedy, W.~Liu, Y.~Jia, P.~Sermanet, S.~E. Reed, D.~Anguelov, D.~Erhan,
  V.~Vanhoucke, and A.~Rabinovich, ``Going deeper with convolutions,'' in
  \emph{Proceedings of the {IEEE} Conference on Computer Vision and Pattern
  Recognition}, 2015, pp. 1--9.

\bibitem{MN:18}
M.~Sandler, A.~G. Howard, M.~Zhu, A.~Zhmoginov, and L.~Chen, ``Mobilenetv2:
  Inverted residuals and linear bottlenecks,'' in \emph{Proceedings of the
  {IEEE} Conference on Computer Vision and Pattern Recognition}, 2018, pp.
  4510--4520.

\bibitem{EN:19}
M.~Tan and Q.~V. Le, ``Efficientnet: Rethinking model scaling for convolutional
  neural networks,'' in \emph{36th International Conference on Machine
  Learning}, 2019, pp. 6105--6114.

\bibitem{CIFAR10:09}
A.~Krizhevsky, ``Learning multiple layers of features from tiny images,'' Tech.
  Rep., 2009.

\bibitem{SVHN:11}
Y.~Netzer, T.~Wang, A.~Coates, A.~Bissacco, B.~Wu, and A.~Y. Ng, ``Reading
  digits in natural images with unsupervised feature learning,'' in \emph{NIPS
  Workshop on Deep Learning and Unsupervised Feature Learning}, 2011.

\bibitem{ISUN:15}
P.~Xu, K.~A. Ehinger, Y.~Zhang, A.~Finkelstein, S.~R. Kulkarni, and J.~Xiao,
  ``Turkergaze: Crowdsourcing saliency with webcam based eye tracking,''
  \emph{CoRR}, vol. abs/1504.06755, 2015.

\bibitem{LSUN:15}
F.~Yu, Y.~Zhang, S.~Song, A.~Seff, and J.~Xiao, ``{LSUN:} construction of a
  large-scale image dataset using deep learning with humans in the loop,''
  \emph{CoRR}, vol. abs/1506.03365, 2015.

\bibitem{IMAGENET:09}
J.~Deng, W.~Dong, R.~Socher, L.~Li, K.~Li, and F.~Li, ``Imagenet: {A}
  large-scale hierarchical image database,'' in \emph{Proceedings of the {IEEE}
  Conference on Computer Vision and Pattern Recognition}, 2009, pp. 248--255.

\bibitem{CLA:15}
Z.~Liu, P.~Luo, X.~Wang, and X.~Tang, ``Deep learning face attributes in the
  wild,'' in \emph{Proceedings of the {IEEE} Conference on Computer Vision and
  Pattern Recognition}, 2015, pp. 3730--3738.

\bibitem{VisDA:17}
X.~Peng, B.~Usman, N.~Kaushik, J.~Hoffman, D.~Wang, and K.~Saenko, ``Visda: The
  visual domain adaptation challenge,'' \emph{CoRR}, vol. abs/1710.06924, 2017.

\bibitem{CAL:06}
G.~Griffin, A.~Holub, and P.~Perona, ``The caltech 256,'' Caltech Technical
  Report, Tech. Rep., 2006.

\bibitem{PASC:17}
D.~Li, Y.~Yang, Y.~Song, and T.~M. Hospedales, ``Deeper, broader and artier
  domain generalization,'' in \emph{Proceedings of the {IEEE} Conference on
  Computer Vision and Pattern Recognition}, 2017, pp. 5543--5551.

\bibitem{COCO:14}
T.~Lin, M.~Maire, S.~J. Belongie, J.~Hays, P.~Perona, D.~Ramanan,
  P.~Doll{\'{a}}r, and C.~L. Zitnick, ``Microsoft coco: Common objects in
  context,'' in \emph{Proceedings of the European Conference on Computer
  Vision}, vol. 8693, 2014, pp. 740--755.

\bibitem{TP:21}
H.~Touvron, M.~Cord, M.~Douze, F.~Massa, A.~Sablayrolles, and H.~J{\'{e}}gou,
  ``Training data-efficient image transformers {\&} distillation through
  attention,'' in \emph{38th International Conference on Machine Learning},
  2021, pp. 10\,347--10\,357.

\bibitem{AUROC:06}
J.~Davis and M.~Goadrich, ``The relationship between precision-recall and {ROC}
  curves,'' in \emph{23th International Conference on Machine Learning}, 2006,
  pp. 233--240.

\bibitem{SNE:08}
L.~van~der Maaten and G.~Hinton, ``Visualizing data using t-sne,''
  \emph{Journal of Machine Learning Research}, vol.~9, no.~86, pp. 2579--2605,
  2008.

\end{thebibliography}


% \newpage
\begin{IEEEbiography}[{\includegraphics[width=1in,height=1.25in,clip,keepaspectratio]{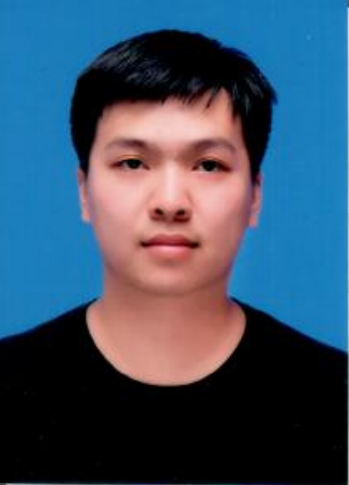}}]{Zhilin Zhao}
received the Ph.D. degree from the University of Technology Sydney in 2022. Prior to that, he received the B.S. and M.S. degrees from the School of Data and Computer Science, Sun Yat-Sen University, China, in 2016 and 2018, respectively. He is currently a Post-Doctoral Fellow at Macquaire University, Australia. His research interests include generalization analysis, dynamic data streams, and out-of-distribution detection.
\end{IEEEbiography}

\begin{IEEEbiography}[{\includegraphics[width=1in,height=1.25in,clip,keepaspectratio]{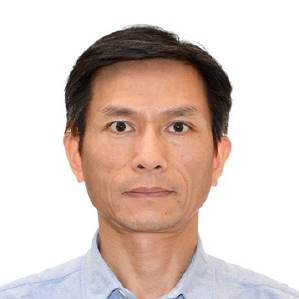}}]{Longbing Cao}(SM'06) received a PhD degree in pattern recognition and intelligent systems at Chinese Academy of Sciences in 2002 and another PhD in computing sciences at University of Technology Sydney in 2005. He is the Distinguished Chair Professor in AI at Macquarie University and an Australian Research Council Future Fellow (professorial level). His research interests include AI and intelligent systems, data science and analytics, machine learning, behavior informatics, and enterprise innovation.
\end{IEEEbiography}

\begin{IEEEbiography}[{\includegraphics[width=1in,height=1.25in,clip,keepaspectratio]{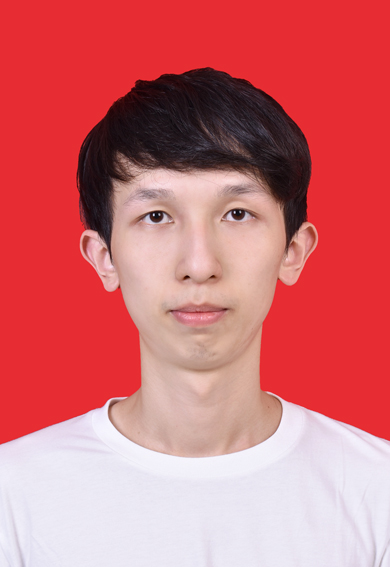}}]{Kun-Yu Lin}
received the B.S. and M.S. degree from the School of Data and Computer Science, Sun Yat-sen University, China, in 2017 and 2019, respectively.
He is currently a PhD student in the School of Computer Science and Engineering, Sun Yat-sen University. His research interests include computer vision and machine learning.
\end{IEEEbiography}


%\vfill


\end{document}